\documentclass[Afour,sageh,times]{nosagej}

\usepackage{moreverb,url}

\usepackage[draft, colorlinks,bookmarksopen,bookmarksnumbered,citecolor=black,urlcolor=black]{hyperref}

\usepackage{graphicx}
\usepackage[noend]{algpseudocode}
\usepackage{color}
\usepackage[dvipsnames]{xcolor}
\usepackage{amsmath,amssymb,amsfonts,dsfont}
\usepackage{bm}
\usepackage{setspace}
\usepackage{multirow}
\usepackage{fancyhdr}
\usepackage{authblk}
\usepackage{caption}
\usepackage{tikz}
\usetikzlibrary{calc}
\usetikzlibrary{decorations.pathreplacing}
\usepackage{gensymb}

\usepackage{algorithm}
\usepackage{algorithmicx}
\usepackage{algpseudocode}

\usepackage[
  separate-uncertainty = true,
  multi-part-units = repeat
]{siunitx}
\usepackage{threeparttable}

\usepackage{renew}      

\usepackage[switch]{lineno} 

\newcommand{\detCost}{L_{\text{det}}}
\newcommand{\loss}[2]{\mathcal{L}_{\text{#1}}^{\text{#2}}}

\newcommand{\gk}[1]{{#1}^{\text{gk}}}
\newcommand{\aux}[1]{{#1}^{\text{cen}}}

\newcommand{\sfigref}[2]{\ref{#1}\,(#2)}
\newcommand{\refsp}[1]{\ref{#1}\,}   

\newcommand{\Real}{\mathbb{R}}

\newcommand{\featSpace}{\mathcal{F}}

\newcommand{\oclass}{\mathcal{C}}
\newcommand{\oel}{c}
\newcommand{\hmap}{{\hat Y}}
\newcommand{\omap}{{\hat O}}
\newcommand{\hmat}{{\hat y}}
\newcommand{\omat}{{\hat o}}

\newcommand{\hmatGT}{{y}}
\newcommand{\omatGT}{{o}}

\newcommand{\smLone}{{L}_1}

\newcommand{\redFact}{R}
\newcommand{\emap}{{\hat E}}
\newcommand{\emat}{{\hat e}}

\newcommand{\pregrasp}{\gamma}
\newcommand{\orient}{\hat \oel}
\newcommand{\point}{p}
\newcommand{\gconf}{s}

\newcommand{\topk}{k}

\newcommand{\threshCenter}{\rho_{\rm cen}}
\newcommand{\threshOrient}{\tau_{\rm orient}}
\newcommand{\threshEmbed}{\rho_{\rm embed}}

\newcommand{\gripRegion}{\mathcal{G}}
\newcommand{\objRegion}{\mathcal{O}}
\newcommand{\Hside}{{\rm H}}

\graphicspath{ {../figs/} }

\newcommand\BibTeX{{\rmfamily B\kern-.05em \textsc{i\kern-.025em b}\kern-.08em
T\kern-.1667em\lower.7ex\hbox{E}\kern-.125emX}}

\begin{document}

\runninghead{Xu et al}

\title{\LARGE \bf
GKNet: grasp keypoint network for grasp candidates detection}

\author{Ruinian Xu, Fu-Jen Chu and Patricio A. Vela\affilnum{1}}

\affiliation{\affilnum{1}Intelligent Vision and Automation Laboratory
(IVALab), School of Electrical and Computer Engineering, Institute for Robotics
and Intelligent Machines; Georgia Institute of Technology, GA, USA.}

\corrauth{Ruinian Xu, Intelligent Vision and Automation Laboratory (IVALab),
Georgia Institute of Technology,
North Ave NW,
Atlanta, GA 30332.}

\email{rnx94@gatech.edu}

\begin{abstract}
Contemporary grasp detection approaches employ deep learning to achieve
robustness to sensor and object model uncertainty.
The two dominant approaches design either grasp-quality scoring or
anchor-based grasp recognition networks.
This paper presents a different approach to grasp detection by treating it
as keypoint detection in image-space.
The deep network detects each grasp candidate as a pair of keypoints,
convertible to the grasp representation 
$g = {\{x, y, w, \theta\}^T}$,
rather than a triplet or quartet of corner points. 
Decreasing the detection difficulty by grouping keypoints into pairs
boosts performance.
To promote capturing dependencies between keypoints,
a non-local module is incorporated into the network design.
A final filtering strategy based on discrete and continuous orientation
prediction removes false correspondences and further improves grasp
detection performance.
GKNet, the approach presented here, achieves a good balance between
accuracy and speed on the Cornell and the abridged Jacquard datasets 
(96.9\% and 98.39\% at 41.67 and 23.26 fps). 
Follow-up experiments on a manipulator evaluate GKNet using 4 types of
grasping experiments reflecting different nuisance sources: 
static grasping, dynamic grasping, grasping at varied camera angles,
and bin picking. 
GKNet outperforms reference baselines in static and dynamic grasping
experiments while showing robustness to varied
camera viewpoints and moderate clutter.
The results confirm the hypothesis that grasp keypoints are an effective
output representation for deep grasp networks that provide robustness
to expected nuisance factors.
\end{abstract}


\maketitle

\section{Introduction}
Robotic manipulation in unstructured and dynamic environments is a
crucial task when deploying robots for daily living and industrial activities.  Grasp detection is an essential sub-problem of robotic manipulation since knowing where and how to grasp objects is a starting point for manipulation.
Real-world use of robot manipulators demands accurate, robust, and efficient or real-time operation. Contemporary solutions aim to exhibit all of these characteristics, however there usually exists a trade-off between them.  
Deep learning based approaches for grasp detection have best balanced these needs to achieve state-of-the-art performance.  
Research for grasping roughly splits into 4 Degree-of-Freedom (DoF) and
6 DoF methods, with 4 DoF methods exhibiting a more favorable balance.
%
%
This paper focuses on 4 DoF grasp recognition such that the output grasp
representation is specified in image coordinates, and converted to a 6 DoF
equivalent through the assumption of top-down grasping.
Image-based, deep learning approaches tend to be convolutional neural network (CNN) approaches 
\citep{redmon2015real, kumra2017robotic, guo2017hybrid, zhou2018fully, chu2018real, asif2019densely}. 
The prevailing neural network approaches divide into one-stage and
two-stage grasp detectors, and grasp quality scoring networks.
One-stage grasp detectors generate grasp predictions with a single deep network, whereas two-stage grasp detectors generate candidate detection regions with one deep network and then evaluate the feature vectors of the candidate regions using a second network to generate grasp hypotheses. 
Compared to one-stage grasp detectors, two-stage grasp detectors
typically achieve better performance by predicting potential grasp regions.
However the incorporation of a grasp region proposal network increases 
inference latency.
One-stage grasp detectors commonly run at real-time speed but with reduced performance.  

Contemporary computer vision research into reducing the performance gap
between one-stage and two-stage detection methods investigates
alternative detection representations within the processing pipeline.
One architecture commonly used is anchor boxes \citep{ren2015faster}; a
series of boxes whose size and aspect ratio lie within a specified
parametric domain. 
Much like for region proposals, a major drawback of the anchor boxes is
the introduction of additional hyperparameters, which require manual
specification to optimize performance. These hyperparameters specify the
total number of anchor boxes and their geometry (such as sizes and aspect ratios). Hyperparameter choices impact detection outcomes.  Another
achitecture is that of keypoints \citep{law2018cornernet, duan2019centernet};
it represents bounding boxes as a set of keypoints. 
Instead of directly regressing parameters for bounding boxes, keypoint
detection approaches perform pixel-level segmentation and yield a set of
keypoint heatmaps. These keypoint maps are then grouped and
post-processed to form the final detections. Keypoint detection methods
are one-stage approaches that achieve competitive performance with
two-stage approaches.

\begin{figure}[t]
    \centering
    \includegraphics[width=0.95\columnwidth,clip=true,trim=0in 0.625in 0in 0.125in]{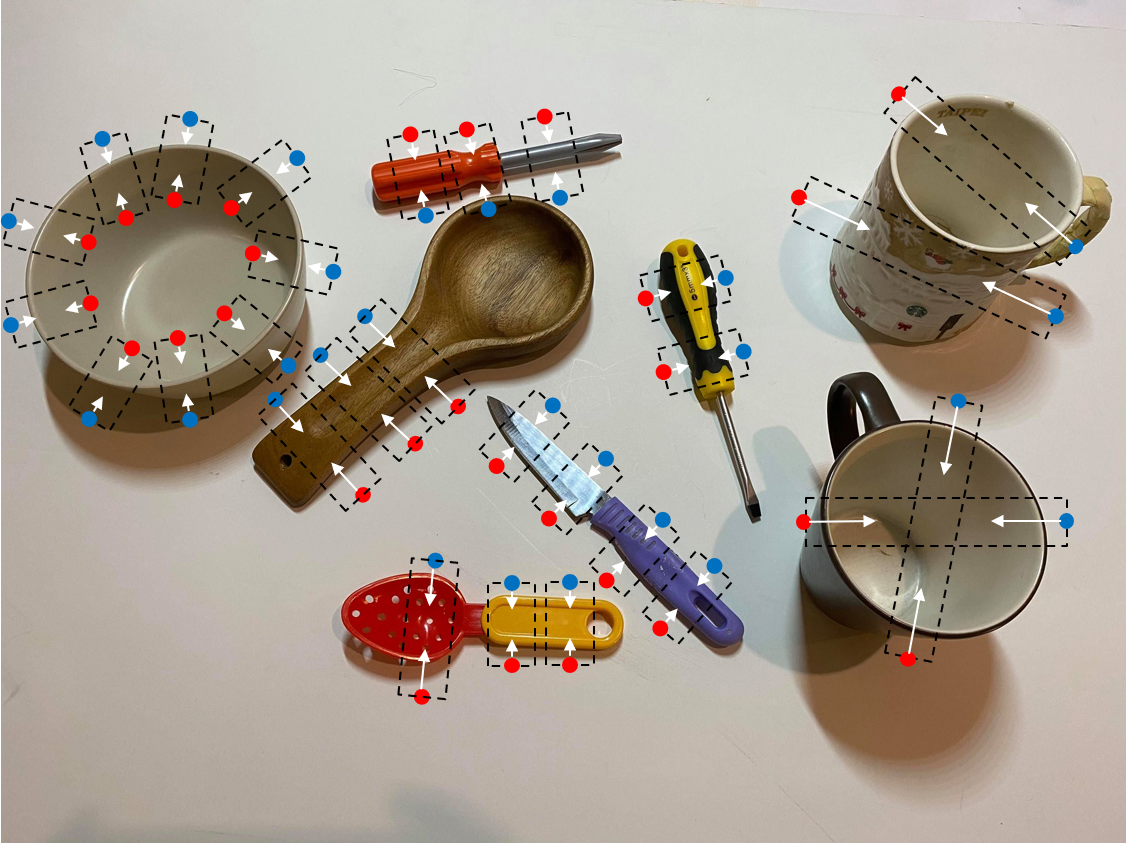} 
    \caption{Illustration of grasp candidate detection with proposed grasp representation
    and method. 
    The red points are right keypoints.
    The blue points are left keypoints. 
    The black boxes correspond to associated grasp bounding boxes for keypoint pairs. 
    For illustration of the grasp, the white arrows indicate the closure
    directions for parallel-jaw grippers. 
    \label{fig_ill}}
\end{figure}

This paper describes a one-stage grasp detector that solves grasp detection as a keypoint detection problem. 
The most general image-based grasp representation,
$g = (x, y, \theta, w, h)^T$, describes an oriented bounding
box.  Though applicable to a variety of
grippers, it is best suited to parallel-plate/jaw gripper types.  
For keypoint architectures, specifying an oriented bounding box
would involve either three or four corner points, whose geometry must
meet orthogonality constraints to properly define a box shape. 
Instead of a fully-specified grasp, the proposed approach represents 
oriented bounding boxes as grasp keypoint pairs: the left-middle
and right-middle points of the grasp bounding box (see Fig.
\ref{fig_ill}). 
Doing so reduces the necessary geometric constraints and also the
difficulty of grouping predicted keypoints to arrive at grasp
candidates, as only pair-wise keypoint matches are needed (versus
triple or quadruple correspondences). The reduction speeds up and
simplifies the grouping process, and provides for additional means to
internally certify correctness of the grasp hypothesis.
Furthermore, 
due to the difficulty of encoding geometric constraints between 
grasp keypoints into the training objective function,
removing redundant components benefits keypoint detection methods, since
even one bad predicted keypoint can significantly change the bounding
box geometry.
A byproduct of the reduced representation is the loss of the 
bounding box length coordinate, $h$. The loss is minimal
since most (parallel-plate) grippers have constant thickness fingertips.
A variable $h$ is rarely implemented. 


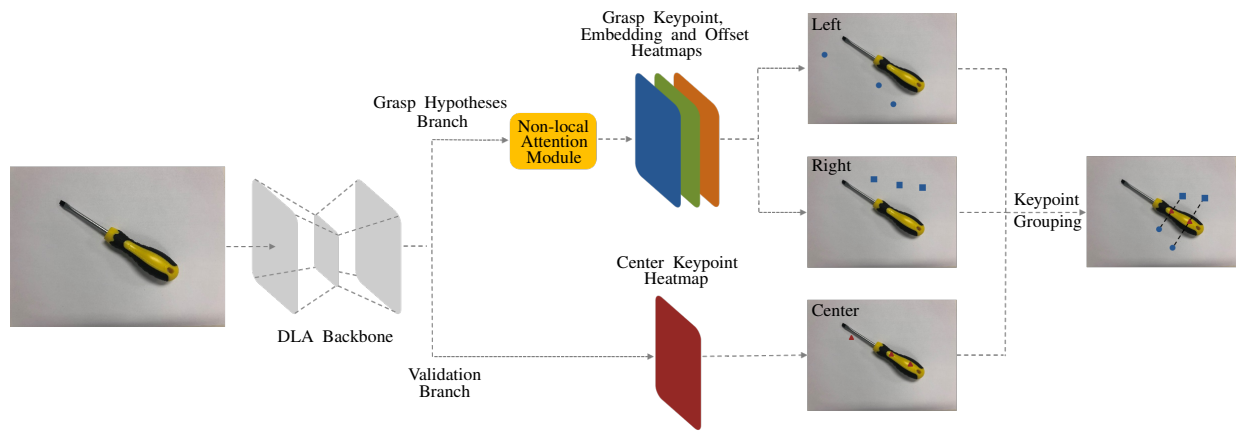
\begin{figure*}[t]
    \centering
    \scalebox{0.95}{
    \begin{tikzpicture} [outer sep=0pt, inner sep=0pt]
    \node[anchor=north west] (a) at (0in,0in) 
    {\includegraphics[width=0.975\textwidth,clip=true,trim=0in 0in 0in 0in]{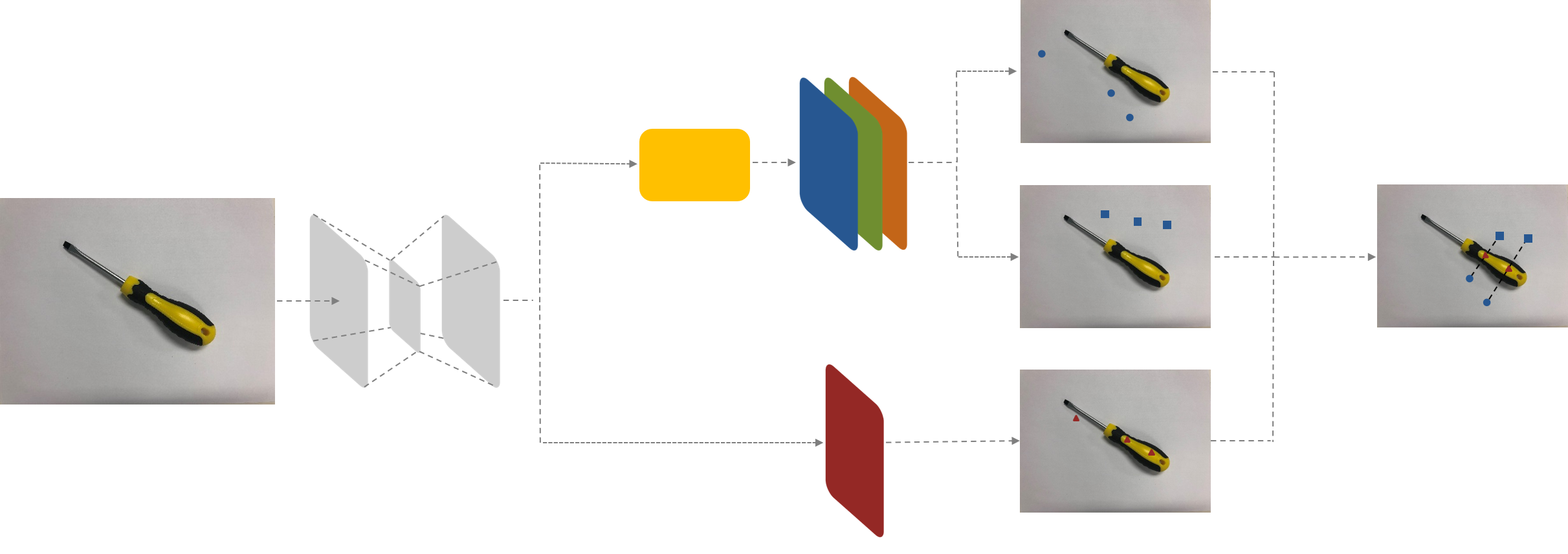}}; 
    \node[text width=2.0cm] at (4.7, -4.5) {\scriptsize DLA Backbone};
	\node[text width=4.0cm] at (13.10, -0.15) {\scriptsize Left};
    \node[text width=4.0cm] at (13.10, -2.15) {\scriptsize Right};
    \node[text width=4.0cm] at (13.10, -4.15) {\scriptsize Center};

    \node[text width=4.0cm] at (7.03, -1.27) {\scriptsize Grasp Hypotheses};
    \node[text width=4.0cm] at (7.63, -1.51) {\scriptsize Branch};
    
    \node[text width=4.0cm] at (7.51, -5.04) {\scriptsize Validation};
    \node[text width=4.0cm] at (7.66, -5.28) {\scriptsize Branch};

	\node[text width=4.0cm] at (9.03, -1.57) {\scriptsize Non-local};
    \node[text width=4.0cm] at (9.07, -1.79) {\scriptsize Attention};
    \node[text width=4.0cm] at (9.15, -2.01) {\scriptsize Module};

    \node[text width=4.0cm] at (10.2, -0.08) {\scriptsize Grasp Keypoint,};
    \node[text width=4.0cm] at (9.9, -0.32) {\scriptsize Embedding and Offset};
    \node[text width=4.0cm] at (10.6, -0.54) {\scriptsize Heatmaps};

    \node[text width=4.0cm] at (10.4, -3.50) {\scriptsize Center Keypoint};
    \node[text width=4.0cm] at (10.78, -3.72) {\scriptsize Heatmap};    
    
    \node[text width=1.0cm] at (14.4, -2.62) {\scriptsize Keypoint};
    \node[text width=1.0cm] at (14.4, -2.92) {\scriptsize Grouping};
    \end{tikzpicture}}
    \vspace*{-1ex}
    \caption{Architecture of GKNet. An hourglass-like backbone 
    network is followed by two prediction branches, one for 
    grasp keypoints and one for a center keypoint. 
    A non-local attention module is inserted between backbone 
    network and the prediction branch for grasp keypoints.
    The pair of \emph{grasp keypoint heatmaps} with corresponding 
    embeddings and offsets and the \emph{center keypoint 
    heatmap} are fed into keypoint grouping algorithm and yield the final grasp detections. Blue dots correspond to left grasp keypoints, blue rectangles correspond to right grasp keypoints and red triangles correspond to center keypoints. 
    \label{fig_netarch}
    }
\end{figure*}

The benefits of a left-middle and right-middle keypoint representation 
translates to neural network processing architecture.
The key-point representation requires less information to synthesize a
valid grasp candidate. Another consequence is that it permits simpler
grasp validation strategies and the incorporation of strategies to
couple the left and right key-point detection processes within the
feature space processing.  The net outcome is to better balance
trade-offs in performance and processing speed.  
The proposed deep network, called \textit{GKNet}, is a one-stage neural
network that maps RGB-D input images to rank-ordered 
${\{x, y, \theta, w\}^T}$ grasp candidates. The GKNet pipeline is depicted
in Fig.~\ref{fig_netarch} and detailed in Section \ref{sec:Approach}.
Internally, the network generates two \emph{grasp keypoint heatmaps},
left and right, along with additional maps to improve keypoint pairing
performance and localization. These maps are the embedding map, offset
map, and {\em center keypoint heatmap}.
Post-processing of the network output groups left and right keypoints
into pairs representing grasp hypotheses based on a series of validation and
filtering conditions using the embedding maps and the center keypoint
heatmap.
%
%
%
%

To quantify the balance between performance and processing speed, GKNet
is compared to baseline algorithms on the purely visual problem of grasp
detection (Section \ref{sec:Vision}\,). 
Relative to the baselines, it achieves good trade-off between
accuracy and speed on the Cornell and Jacquard Datasets \citep{cornell2013, depierre2018jacquard}.  It lies at the extremal corner of the convex hull of all methods in the accuracy/speed plane.  GKNet achieves competitive performance with two-stage approaches. 
To quantify performance when deployed on an actual manipulator, GKNet is
integrated into a manipulation pipeline and tested on four physical
grasping experiments:
(a) static grasping, 
(b) dynamic grasping, 
(c) grasping at varied camera viewpoints and 
(d) bin picking. 
These tests, described in Section \ref{sec:PE}, evaluate the
system with respect to accuracy, speed, and robustness. 
Reported experiment outcomes in Section \ref{secResGrasp} include
comparative discussions and analysis.
Performance outcomes across the suite of benchmarks for static grasping, 
which include some existing one-stage and two-stage grasp detectors,
demonstrate that GKNet is capable of predicting more accurate grasping
candidates.
Dynamic grasping experiment results show that GKNet provides reliable
grasp predictions in real-time. 
Grasping at varied camera poses and bin picking experiments illustrate
the robustness of GKNet under different camera views and in cluttered
environments scenarios, which are not included in the training dataset.
The consistent performance of GKNet across these experiments provides
evidence in favor of keypoint representations for grasp recognition.

\section{Related Work}
Research emphasis on robotic grasp detection has shifted over the past two
decades from analytical methods to empirical methods.  The review papers
\citep{bicchi2000robotic, sahbani2012overview, shimoga1996robot} describe
conventional analytical methods. They perform grasp candidate
analysis based on prior knowledge, like object geometry or pose, 
grasp closure and force analytics, and/or physical models. 
The major drawback lies in the assumption that the object parameters
informing the grasp analysis are known, which is hard to satisfy 
for general, real-world manipulation. 
Empirical methods, also known as learning-based or data-driven methods,
perform grasp detection using functional models learned from 
available real or synthetic grasp results and their input signals
(usually imagery).  When given diverse inputs, empirical methods
generalize to novel cases.
This section reviews existing grasp detection approaches with an
emphasis on the empirical methods.

Additionally, the focus will be on parallel-jaw grippers or their
equivalents given that a large class of objects can be grasped from two
opposing contact regions. Where this class of grippers can fail is with
non-smooth or asymmetric grasping surfaces.
Multi-fingered grippers are capable of performing more complex grasp
configurations 
\citep{lu2020planning, kappler2015leveraging, saxena2008learning, 
varley2015generating, veres2017modeling, zhou20176dof}. 
Alternatively, suction gripper platforms provide a different set of
grasping affordances for grasping diverse categories of objects based on
a single contact region
\citep{eppner2016lessons,mahler2018dex,shao2019suction}
which has led to their wide deployment in warehouse or manufacturing
settings. Their common use by industry is backed by recent research
strategies that lead to suction grippers being favored over standard
grippers due to their grasping performance advantage, when employing a
bi-manual suction and parallel-plate gripper robot configuration 
\citep{mahler2019learning,zeng2018robotic,yen2020learning,fujita2020important}.
Though highly capable when used in structured settings like warehouses or
manufacturing, suction grasping limits the target objects to those with
suction graspable surface geometry. 
This work aims to improve parallel-plate styled gripper grasp success
and lower the performance gap for suction grippers while improving on
use cases and scenarios where the latter are not used/available.

Continuing, early empirical methods for transferring prior knowledge of
annotation grasps to novel objects employed manually designed feature
representations. The idea of decomposing grasping into the
identification of where and how to grasp (location and end-effector
pose) and how well the location will grasp (quality) emerged as an
important design template
\citep{kamon1996learning}.
Such a decoupled learning process meant that location parameters were
object-agnostic and could be applied to new objects.
Likewise, the notion that learning might be best performed through
demonstration instead of annotation has been a consistent thread of
research activity.
Demonstration can lead to direct mappings from the demonstrated human
grasp postures to robot grasps \citep{ekvall2007learning}, or
indirect mappings that transform grasp contact points for objects in the
human demonstration database to novel or known target objects
\citep{hsiao2006imitation}. 

Since then, advances in compute performance and machine learning, plus
the generation of larger annotated datasets, contributed to increased
study into supervised learning techniques. Employing learning approaches
sourced from the computer vision community, methods for inferring grasp
directly from visual data, RGB or RGB-D, quickly came to prominence
\citep{saxena2008robotic,le2010learning}. 
Working in image space required defining grasps in terms of oriented
rectangular grasp regions that were then mapped to an actual 3D grasp
configuration and pre-grasp pose, then planned for execution
\citep{jiang2011efficient}. 
Some methods chose instead to operate on the point clouds and surface
representations \citep{pelossof2004svm,rao2010grasping}. The top
performing learning-based methods demonstrated robust grasping by
manipulators with average success rates in the range of 85-90\% for
small sets of novel objects.
Likewise, simulation methods emerged as a means to rapidly collect
training data for well defined feature spaces \citep{pelossof2004svm}.
In some instances reinforcement learning over simple feature spaces was
used to identify quality grasps 
\citep{piater2002learning,coelho2001developing}. 
These early studies informed the conceptual framing of grasping, while
later work has improved the outcomes through the use of deep learning
methods with more robust inference properties. The remainder of this
section reviews grasping based on deep neural network implementations.

\subsection{Deep Learning Methods}

The asserted benefit of deep learning is that the loss functions
implicitly shape the internal feature space and free the designer from
having to explicitly encode the input to feature space mapping. When
combined with large datasets, presumably the internal feature space
supports generalization to novel inputs or robustness to input variation.
Deep networks have been used for grasp recognition, for grasp quality
scoring, and within grasp policy reinforcement learning strategies.
These are all reviewed below.

\subsubsection{Grasp Quality Scoring.}
As an extension of their analytical predecessors, grasp quality scoring
networks use deep networks to evaluate sampled grasp candidates and
create a ranked ordering based on hypothesized grasp success scores.
Instead of end-to-end mapping images to grasps or actions, these methods
consider robotic grasping within a pipeline where earlier steps involve
grasp sampling and later steps involve grasp planning algorithms. 
Treating grasp detection as a binary classification problem of graspable
or not graspable is the most straightforward implementation
\citep{gualtieri2016high}. 
This idea was extended in \cite{pinto2016supersizing} with the inclusion
of orientation dependent scoring implemented as an 18-way classifier
over discretized orientation classes.
Since grasp samples may not densely nor sufficiently sample from the
feasible grasp space, there is some advantage to using deep learning to
first hypothesize candidate grasps \citep{lenz2015deep}. In that case, a
second network will then evaluate the detected candidates and provide a
ranked ordering.
When trained with binary labels from annotated grasp datasets, there
will be an upper limit on performance since true grasp quality scores
are not available.
Based on the assumption that not all grasps have the same grasp success
probability, extending binary labels to ordered labels in the
annotations better informs the neural network learning process with
regards to ranking grasp candidates \citep{johns2016deep}.
Once actual scores are available, the function output can be modified to
perform continuous regression and avoid the quantization effects of
classifier-based network designs \citep{mahler2017dex}.
Since doing so requires more annotations, \cite{mahler2017dex} included
the creation of Dex-Net 2.0, a dataset with 6.7 million points clouds
each of which is tagged with grasps and their scores as generated by
analytic grasp metrics. Continuing further, imitation learning
employing simulated grasp scenarios can provide even richer training
data for grasp scoring and annotation 
\citep{mahler2017learning, zeng2018learning, yen2020learning}. 
Furthermore, the simulated scenarios can be custom tailored to specific
manipulation tasks, such as bin picking \citep{mahler2017learning}.
Besides learning a grasp quality function, some works employ other
features for representing the graspability of grasp candidates. 
For example, the distance between the manipulator and the nearest
viable grasp after executing the given action 
provides an indirect probability that the manipulator achieves the graspable
configuration \citep{viereck2017learning}.

\subsubsection{2D Grasp Prediction.}
A sub-class of grasp detection or grasp recognition methods map the
input images to a set of 2D oriented bounding boxes as the grasp
representation, with the premise that follow-up processing of the grasp
region will recover the actual 3D grasping approach to apply. The
assumption of a top-down view with top-down grasping provides a direct
conversion of the grasp representation to the 6 DoF grasp to execute.
These grasp recognition strategies decompose into one-stage grasp
detectors, two-stage grasp detectors, and lastly implementations that
combine traditional image and point-cloud processing pipelines and
manually engineered features with convolutional neural networks
\citep{asif2017rgb}. 
The principle idea behind this last category is that some critical
information is best computed and provided directly, as opposed to
expecting the learning process to implicitly recover and encode the
information within its convolutional structure and internal feature
space.  While annotation demands are reduced and performance is strong
for the multi-pronged approach, contemporary research aims to simplify
the processing pipeline while achieving high grasp success rates and
high processing rates.
Generally one-stage methods use a single neural network to directly
regress single or multiple grasp candidates from an image
\citep{redmon2015real, wang2016robot, watson2017real, 
kumra2017robotic}. 
The most commonly used grasp representation 
is $(x, y, \theta, h, w) \in  SE(2) \times \Real^2$ 
\citep{redmon2015real, watson2017real, kumra2017robotic}. 
Some variations add extra or remove parameters such
as the 4-D grasp representation consisting of center location,
opening length and orientation angle \citep{wang2016robot}.
Although these approaches provided relatively high (around 90\%) grasp
detection rate on the Cornell Dataset \citep{cornell2013}, performance is
lower than two-stage detectors.

An essential technique introduced to help one-stage grasp detectors
break the performance ceiling was that of anchor boxes.  A series of
anchor boxes with different scales and aspect ratios predicts potential
grasp candidates over the input image without the extra process of
scanning the image with a sliding window technique. Applying the
technique to grasping does reduce the performance gap somewhat, and can
also incorporate tactile information to improve the grasp detection
process \citep{guo2017hybrid}. 
Extending to the case of oriented anchor boxes led to one-stage methods
on par with two-stage methods, but with equivalent processing speed
even though the one-stage method should be faster \citep{zhou2018fully}. 
Abstracting the concept of anchor boxes to regions leads to networks
that perform pixel-wise predictions over image regions.
Predicting pixel-wise grasp hypotheses
removes the need to define anchor box hyper-parameters
\citep{morrison2019learning,satish2019policy}.  Both implementatios
are grasp quality networks that also output grasp geometry parameters
such as the rotation angle and opening \citep{morrison2019learning},
or the 3-D position and planar orientation \citep{satish2019policy}.
While using anchor boxes improves the detection accuracy, they
also introduce additional hyperparamters that requires manual
specification to optimize performance.

Compared to one-stage approaches, two-stage grasp detectors require an
extra region proposal network (RPN) for generating regions of interest
(RoIs). A series of RoIs generated in the first stages are used to crop
corresponding features over the entire image for identifying grasp
candidates in the second stage. 
Evaluating all potential grasps, RPN helps locate coarse regions that
may contain potential grasps. Furthermore, the two-stage mechanism
achieves strong performance by first coarsely searching and then
refining detections \citep{chu2018real}. 
Based on the fact that classification-, regression- and
segmentation-based grasp detection methods have their own advantages and
drawbacks, some researchers try to fuse features captured at different
levels to improve on grasp recognition. The resulting network is a
multi-branch deep neural network which predicts grasp candidates at
global, regional, and pixel-levels \cite{asif2019densely}.  
Two-stage methods achieve good performance on grasp detection, but high
inference latency may limit their translation to real-world applications.

\subsubsection{Reinforcement Learning.}
Reinforcement learning (RL) solves robotic grasping by learning end-to-end
robot arm motion control policies from image inputs in a manner that
couples the learning process with the experiential training data. 
RL is effective when ground truth annotations are difficult to come by.
Predicted motions can be represented in joint-space \citep{levine2016end} 
or Cartesian task-space \citep{levine2016learning}.
Research emphasis on this front tends to focus on 
manipulation or robust picking strategies targeted to specific tasks
and working environments. 
For densely cluttered environments, grasp success rate is highly affected
by object stacking configurations.  Rearranging objects prior to grasping
has value.  \cite{zeng2018learning} added pushing to the action-space
with the aim of learning synergies between pushing and grasping actions. 
Fusing grasp measurement signals, such as the pressure measured after
performing a grasp, helps the robot determine re-grasp strategies.
\cite{calandra2018more} combined tactile information with visual
observation to learn an evaluation function that maps the current
visuo-tactile observations and a candidate action to its probability of
success.
End-to-end learning approaches generate a locally optimal action
response by end-to-end mapping observations to robotic motions but their
application for real-world tasks remains uncertain given that a policy's
robustness to configuration changes is not fully known. Better
generalization capabilities for grasping have been shown using other
batch-based learning methods.  For example, more generalizable policy
learning has also been achieved through the use of imitation learning
(IL) strategies in the training process
\citep{mahler2017learning,mahler2019learning,satish2019policy}.  where
IL has strong connections to RL.

\subsubsection{3D Grasp Prediction.}
Contemporary grasp recognition methods with high success rates benefit
from premises that favor manipulation using image-based grasp
representation. For more general 3D grasping, the full $SE(3)$ grasp
pose should be output.
The less general approaches employ object shape recognition
strategies, where the recognized instances provides strong grasp priors
from the object geometry or known grasp strategies relative to the
object's pose. Object recognition methods use the known object category
and its predicted pose to inform grasping \citep{Tremblay2019deep}, 
or use object belief to score grasp quality \citep{mahler2016dex}.

Object-agnostic solutions lie in two categories. 
The first category mainly employs geometric information to approximate 
the object or robot gripper for 3-D grasp detection.
Based on prior object knowledge, model-based methods approximate
objects with box-based shapes \citep{huebner2008selection} or primitive
surface shapes \citep{lin2020using} for candidate grasp generation.
Once the parametric shape is recognized, extracted, and estimated, the
grasps naturally follow.
Early model-free methods \citep{bohg2010learning} utilized edge features of
objects to compute shape context and extract grasp points.
Localized antipodal geometry provides an effective means to identify
good candidate grasps \citep{ten2018using, zapata2017using}.
Instead of estimating object geometry, other works focused on the hand
contact model for multi-fingered grippers and its role in supporting
grasp recognition \citep{sorour2019grasping, kopicki2016one}.
\cite{sorour2019grasping} approximated the gripper model by a set of
ellipsoids and generated valid grasps via the computed object pose.
\cite{kopicki2016one} modeled the contact of hand configurations and
surface features using a kernel-density representation, which was used
to score grasp candidates.

The second category involves methods that process 3D point cloud
information in either a discriminative or a generative manner. 
Discriminative methods \citep{ten2017grasp, liang2019pointnetgpd} focus on
evaluating or scoring grasp hypotheses sampled by a given grasp sampler.
Rather than just predict grasp scores, \cite{lou2020learning} proposed
to further learn grasp pose reachability.
Generative methods \cite{qin2020s4g, ni2020pointnet, wu2020grasp}
performed direct regression of the hypothesized 6-DoF grasp configuration.
By introducing post grasp refinement, \cite{mousavian20196, zhao2020regnet} 
forms a closed-loop grasp detection to improve detection accuracy.
For addressing 6-DoF grasping detection in complex cluttered scenarios,
\cite{murali20206} proposed to also predict the collision score for grasp
candidates.
Following the idea of learning the offset, \cite{jeng2020coarse} proposed
a coarse-to-fine representation for 6-DoF grasps.
To reduce 6-Dof grasp to low-dimensional representation,
\cite{sundermeyer2021contact} proposed to project 6-DoF grasps to 4-DoF grasp
representation conditioned on a known contact point. 
The grasp representation is composed of 3-DoF grasp rotation and grasp width. 
It exploits grasp contact geometry to define specific axes capable of
recovering the entire grasp geometry from 4 parameters for a given
contact point. The approach here employs a similar concept but operates
in image-space instead of using a reconstructed point cloud in the
vicinity of the candidate grasp.  Prior information applied to the 2D
image-based grasp geometry reconstructs the 3D grasp geometry.

\section{Problem Statement}
\newcommand{\kpG}{g_{\text{kp}}}
\newcommand{\rlm}{r_{\text{lm}}}
\newcommand{\rrm}{r_{\text{rm}}}
\newcommand{\clm}[1]{{#1}_{\text{lm}}}
\newcommand{\crm}[1]{{#1}_{\text{rm}}}

This paper explores the impact of the grasp representation on
performance for image-based 2D grasp recognition deep networks, and its
impact on the design of the deep network.  The grasp recognition problem
of interest involves using a color and depth image pair (i.e., RGB-D
image) as input to provide as output potential grasp candidates of the
form $g = (x, y, \theta, w, h)^T$. 
The deep network output used to obtain $g$ will be that of grasp
keypoint pairs. These pairs define the left-middle and right-middle
points of a keypoint grasp
\begin{equation} \label{eqn_keypoint_representation}
  \kpG = [\rlm^T, \rrm^T]^T = [\clm x, \clm y, \crm x, \crm y]^T
\end{equation}
in image coordinates, where the subscript $\clm \cdot$ stands for
left-middle and $\crm \cdot$ for right-middle.
For grippers with constant finger widths, the keypoint grasp
representation is a simpler, alternative method for representing grasp
candidates. It removes from consideration the fixed parameter, $h$, 
of the bounding box as it is not commonly adjusted in typical
scenarios.  

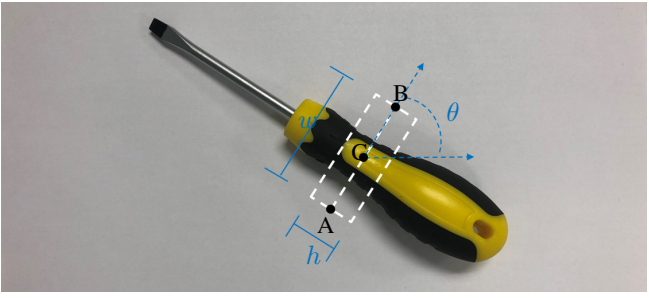
\begin{figure}[t]
  \centering
  \begin{tikzpicture} [outer sep=0pt, inner sep=0pt]
  \node[anchor=north west] (a) at (0in,0in) 
    {\includegraphics[width=\columnwidth,clip=true,trim=0in 1.25in 0in 2.5cm]{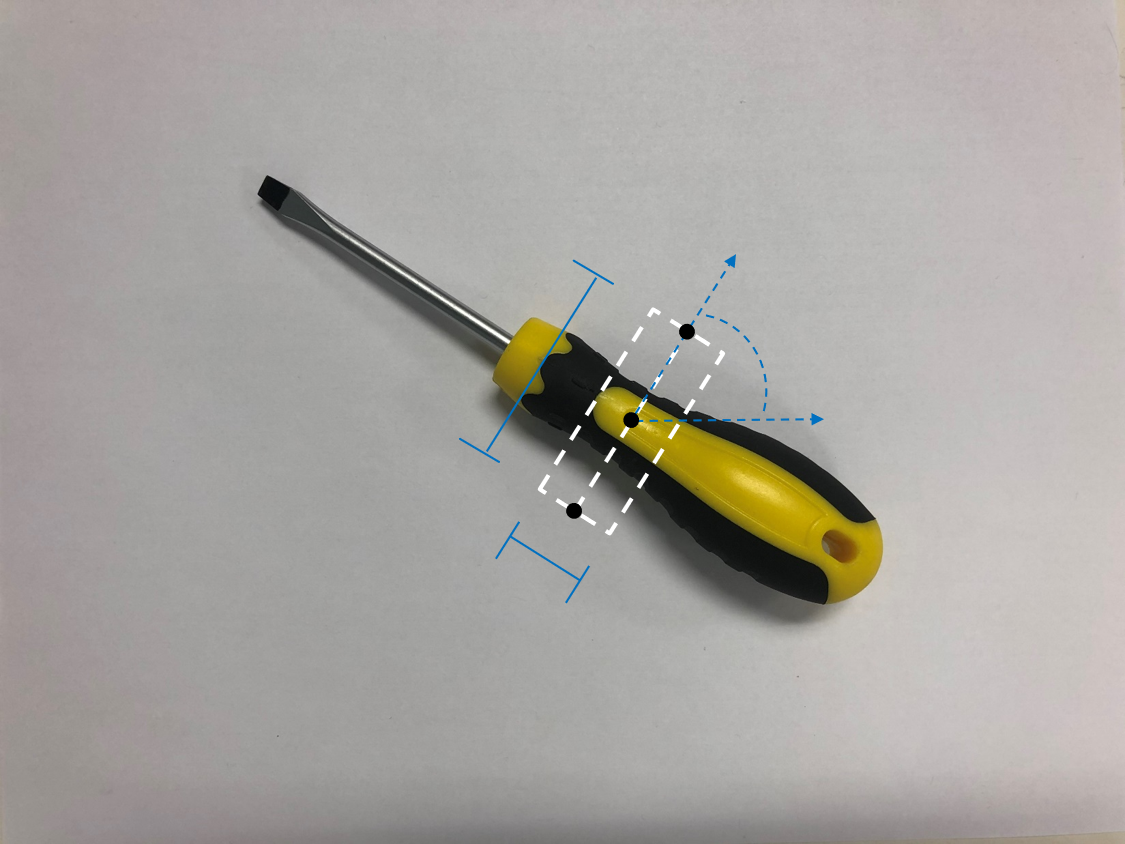}}; 
  \node[text width=1cm] at (4.4cm, -1.6cm) 
    {\textcolor[rgb]{0,0.439,0.753}{$w$}};
  \node[text width=1cm] at (4.5cm, -3.35cm) 
    {\textcolor[rgb]{0,0.439,0.753}{$h$}};
  \node[text width=1cm] at (6.35cm, -1.45cm) 
    {\textcolor[rgb]{0,0.439,0.753}{$\theta$}};
  \node[text width=0.5cm] at (4.4cm, -2.95cm) {\small A};
  \node[text width=0.5cm] at (5.4cm, -1.2cm)  {\small B};
  \node[text width=0.5cm] at (4.85cm, -1.99cm) {\small C};
  \end{tikzpicture}
  \caption{
  Points A and B are two grasp keypoints. Point C is the center
  keypoint. The angle $\theta$ is the continuous rotation angle between
  the line segment $\overline{\textrm{AB}}$ and the x-axis. 
  The width $w$ of the bounding box is recovered from the length of the
  segment $\overline{\textrm{AB}}$. 
  The height $h$ of the bounding box isn't implemented in the proposed
  representation.\label{fig_grasp_rep}}
\end{figure}

After identifying a grasp keypoint pair, as shown in
Fig.~\ref{fig_grasp_rep} with A and B denoting the two keypoints,
conversion of the pair from keypoint grasp coordinates to more standard
grasp coordinates 
\begin{equation} \label{eqn_grasp_representation}
  g = {(x, y, \theta, w)^T},
\end{equation}
is straightforward. In particular, 
$r = (x,y)^T = \frac 12 (\rlm + \rrm)$, 
$\theta = \angle(\rrm - \rlm)$, and $w = \left\|\rlm - \rrm \right\|_{2}$.
Our aim is to show that encoding the image-based, {\em
left-middle and right-middle} representation within a deep network
improves grasp prediction performance and robustness with fast
processing rates, when it is used to solve grasping as a perception
problem and to execute the grasps on robotic manipulators.

\section{Approach \label{sec:Approach}}
This section describes the grasp keypoint network, GKNet, as motivated
by recent one-stage keypoint detection methods such as CornerNet
\citep{law2018cornernet} and CenterNet \citep{duan2019centernet}.
The overall neural network architecture of GKNet is depicted in 
Fig. \ref{fig_netarch}.  The backbone deep network is an hourglass type,
which has an encoder-decoder structure with iterative and
hierarchical skip connections (not depicted in the figure).
There are two branches in the network, one to generate grasp hypotheses
(top branch) and one to validate them (bottom branch).  
The grasp hypothesis branch produces two \emph{grasp keypoint heatmaps}: 
heatmaps for the left-middle and right-middle grasping points.
Each \emph{grasp keypoint heatmap} has additional data output
along with it (e.g., embedding and offset maps) thatserves to refine
and constrain the final grasp keypoint options.
The validation branch generates one \emph{center keypoint heatmap}
trained to output the center keypoint (i.e., the midpoint between the 
left-middle and right-middle keypoints). 

When implemented as a grasp recognition procedure, the processing flow
first uses a neural network trained to predict sets of left-middle,
right-middle, and center keypoints over the input image domain (Section
\subsecref{kpDetect}). Additional data output for these keypoints includes
embedding values, which support the second step of grouping the
left-middle and right-middle keypoints based on embedding value
similarity (Section \subsecref{grouping}).  The identified keypoint
groups are estimated to be valid based on the \emph{center keypoint
heatmap} and a series of conditions for their center keypoints.
An \textit{orientation filter} further refines the final grasp candidate
set (Section \ref{secOrFilter}). Section \subsecref{GKNet} covers
additional details of this process.

\subsection{Keypoint Detection\subseclabel{kpDetect}}

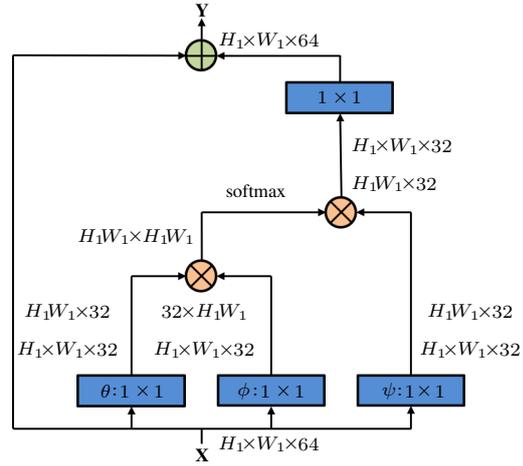
\begin{figure}[t]
    \centering
    \begin{tikzpicture} [outer sep=0pt, inner sep=0pt]
    \node[anchor=north west, xshift=0cm] (a) at (1.5cm,0in) 
    {\includegraphics[width=0.7\columnwidth,clip=true,trim=0in 0in 0in 0in]{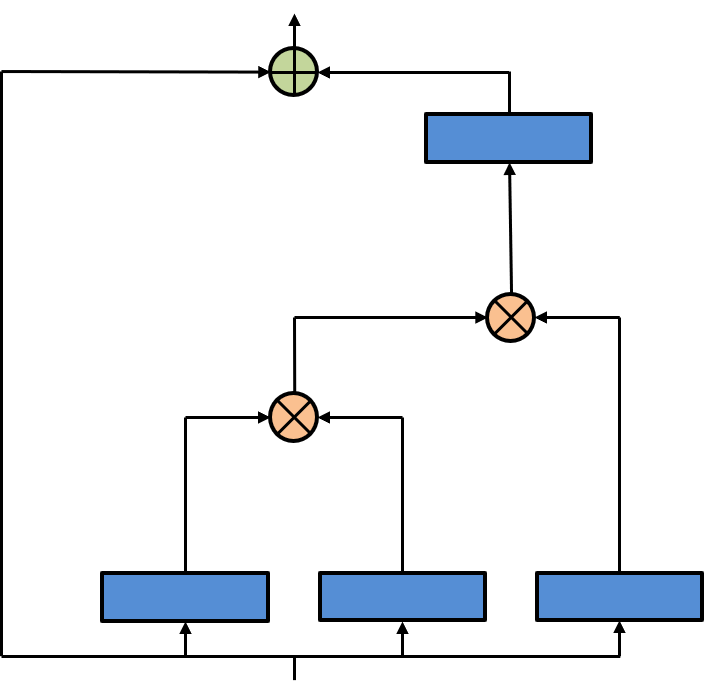}}; 

  	\node[text width=4.0cm] at (5.90, -5.9) {\scriptsize \textbf{X}};
  	
  	\node[text width=4.0cm] at (5.90, -0.0) {\scriptsize \textbf{Y}};
  	
	\node[text width=4.0cm] at (6.20, -5.75) {\scriptsize $H_1$};
    \node[text width=4.0cm] at (6.50, -5.75) {\scriptsize $\times$};
    \node[text width=4.0cm] at (6.70, -5.75) {\scriptsize $W_1$};
    \node[text width=4.0cm] at (7.05, -5.75) {\scriptsize $\times$};
    \node[text width=4.0cm] at (7.25, -5.75) {\scriptsize $64$};  	
  	
    \node[text width=4.0cm] at (3.55, -4.5) {\scriptsize $H_1$};
    \node[text width=4.0cm] at (3.85, -4.5) {\scriptsize $\times$};
    \node[text width=4.0cm] at (4.05, -4.5) {\scriptsize $W_1$};
    \node[text width=4.0cm] at (4.40, -4.5) {\scriptsize $\times$};
    \node[text width=4.0cm] at (4.60, -4.5) {\scriptsize $32$};
    
    \node[text width=4.0cm] at (3.65, -4.0) {\scriptsize $H_1$};
    \node[text width=4.0cm] at (3.95, -4.0) {\scriptsize $W_1$};
    \node[text width=4.0cm] at (4.30, -4.0) {\scriptsize $\times$};
    \node[text width=4.0cm] at (4.50, -4.0) {\scriptsize $32$};
    
    \node[text width=4.0cm] at (5.35, -4.5) {\scriptsize $H_1$};
    \node[text width=4.0cm] at (5.65, -4.5) {\scriptsize $\times$};
    \node[text width=4.0cm] at (5.85, -4.5) {\scriptsize $W_1$};
    \node[text width=4.0cm] at (6.20, -4.5) {\scriptsize $\times$};
    \node[text width=4.0cm] at (6.40, -4.5) {\scriptsize $32$};
    
    \node[text width=4.0cm] at (5.45, -4.0) {\scriptsize $32$};
    \node[text width=4.0cm] at (5.70, -4.0) {\scriptsize $\times$};
    \node[text width=4.0cm] at (5.90, -4.0) {\scriptsize $H_1$};
    \node[text width=4.0cm] at (6.20, -4.0) {\scriptsize $W_1$};
    
    \node[text width=4.0cm] at (8.85, -4.5) {\scriptsize $H_1$};
    \node[text width=4.0cm] at (9.15, -4.5) {\scriptsize $\times$};
    \node[text width=4.0cm] at (9.35, -4.5) {\scriptsize $W_1$};
    \node[text width=4.0cm] at (9.70, -4.5) {\scriptsize $\times$};
    \node[text width=4.0cm] at (9.90, -4.5) {\scriptsize $32$};
    
    \node[text width=4.0cm] at (8.95, -4.0) {\scriptsize $H_1$};
    \node[text width=4.0cm] at (9.25, -4.0) {\scriptsize $W_1$};
    \node[text width=4.0cm] at (9.60, -4.0) {\scriptsize $\times$};
    \node[text width=4.0cm] at (9.80, -4.0) {\scriptsize $32$};
    
    \node[text width=4.0cm] at (6.30, -2.4) {\scriptsize \text{softmax}};
    
    \node[text width=4.0cm] at (4.35, -3.0) {\scriptsize $H_1$};
    \node[text width=4.0cm] at (4.65, -3.0) {\scriptsize $W_1$};
    \node[text width=4.0cm] at (5.00, -3.0) {\scriptsize $\times$};
    \node[text width=4.0cm] at (5.20, -3.0) {\scriptsize $H_1$};
    \node[text width=4.0cm] at (5.50, -3.0) {\scriptsize $W_1$};
    
    \node[text width=4.0cm] at (7.95, -2.3) {\scriptsize $H_1$};
    \node[text width=4.0cm] at (8.25, -2.3) {\scriptsize $W_1$};
    \node[text width=4.0cm] at (8.60, -2.3) {\scriptsize $\times$};
    \node[text width=4.0cm] at (8.80, -2.3) {\scriptsize $32$};
    
    \node[text width=4.0cm] at (7.95, -1.8) {\scriptsize $H_1$};
    \node[text width=4.0cm] at (8.25, -1.8) {\scriptsize $\times$};
    \node[text width=4.0cm] at (8.45, -1.8) {\scriptsize $W_1$};
    \node[text width=4.0cm] at (8.80, -1.8) {\scriptsize $\times$};
    \node[text width=4.0cm] at (9.00, -1.8) {\scriptsize $32$};

	\node[text width=4.0cm] at (6.20, -0.4) {\scriptsize $H_1$};
    \node[text width=4.0cm] at (6.50, -0.4) {\scriptsize $\times$};
    \node[text width=4.0cm] at (6.70, -0.4) {\scriptsize $W_1$};
    \node[text width=4.0cm] at (7.05, -0.4) {\scriptsize $\times$};
    \node[text width=4.0cm] at (7.25, -0.4) {\scriptsize $64$}; 
    
    \node[text width=4.0cm] at (4.65, -5.05) {\scriptsize $\theta$};
    \node[text width=4.0cm] at (4.80, -5.05) {\scriptsize $:$};
    \node[text width=4.0cm] at (4.90, -5.05) {\scriptsize $1$};
    \node[text width=4.0cm] at (5.10, -5.05) {\scriptsize $\times$};
    \node[text width=4.0cm] at (5.35, -5.05) {\scriptsize $1$}; 
    
    \node[text width=4.0cm] at (6.45, -5.05) {\scriptsize $\phi$};
    \node[text width=4.0cm] at (6.65, -5.05) {\scriptsize $:$};
    \node[text width=4.0cm] at (6.75, -5.05) {\scriptsize $1$};
    \node[text width=4.0cm] at (6.95, -5.05) {\scriptsize $\times$};
    \node[text width=4.0cm] at (7.20, -5.05) {\scriptsize $1$}; 

	\node[text width=4.0cm] at (8.35, -5.05) {\scriptsize $\psi$};
    \node[text width=4.0cm] at (8.55, -5.05) {\scriptsize $:$};
    \node[text width=4.0cm] at (8.65, -5.05) {\scriptsize $1$};
    \node[text width=4.0cm] at (8.80, -5.05) {\scriptsize $\times$};
    \node[text width=4.0cm] at (9.05, -5.05) {\scriptsize $1$};

	\node[text width=4.0cm] at (7.50, -1.16) {\scriptsize $1$};
    \node[text width=4.0cm] at (7.70, -1.16) {\scriptsize $\times$};
    \node[text width=4.0cm] at (7.95, -1.16) {\scriptsize $1$};

    \end{tikzpicture}
    \caption{The non-local module architecture. The feature maps
    are represent by their shapes, e.g., $H_1 \times W_1 \times 1024$ where
    $H_1$, $W_1$ and $1024$ are the height, width and channels. "$\otimes$"
    denotes the matrix multiplication, and "$\bigoplus$" denotes the element-wise
    addition. The blue boxes represent $1 \times 1$ convolutions.
    \label{fig_nonlocal}
    }
\end{figure}

\subsubsection{The grasp hypotheses branch.}
To support the generation of grasp hypotheses, the 
\emph{grasp keypoint heatmaps} are defined over image coordinates and a
quantized set of grasp orientation classes. Given the
symmetry of grasping with parallel plate type grippers, the relevant
domain for the grasp angles lies within a 180 degree range: 
$[-90,90]$ with orientations outside of the domain mapped into it
through the modulo operation.
Denote by $\oclass$ the set of orientation classes.
Each class element $\oel \in \oclass$ represents a 
quantization interval. 
The grasp keypoint branch generates two heatmaps
$\gk \hmap_L, \gk \hmap_R : \oclass \times \Real^2 \rightarrow \Real$ 
stored as matrices 
$\gk \hmap_L,\gk \hmap_R \in  \mathbb{R}^{\left|\oclass\right| \times H \times W}$ 
for the left-middle and right-middle keypoints, where $H$ and $W$ are
the height and width dimensions of the heatmaps.
Each coordinate location indexing into a \emph{grasp keypoint heatmap} 
represents the probability distribution over the orientation classes 
$\oel \in \oclass$. 

Because the left-middle and right-middle keypoints have a geometric
relationship relative to each other and to the graspable region, these
heatmaps should be correlated or coupled to implicitly capture 
important relationships.  
Per Fig. \ref{fig_netarch}, a non-local module is inserted in the grasp
keypoint branch, between the backbone network output and the grasp
keypoint heatmap generator input.
Non-local modules provide simple and efficient mechanisms to capture
long-range spatial dependencies for deep CNNs, dependencies that may not
be captured through the standard convolution and pooling operations due
to their locality properties \citep{wang2018non}. 
The module contributes to learning larger scale spatial dependencies
that improve the performance of downstream processing.
As shown in Figure \ref{fig_nonlocal}, the input and output dimensions
are the same. The intermediate processing within the module generates a
response output at a given location as a weighted sum of the features at
all locations of the input feature map. The training process couples the
outputs at given locations to feature vectors that lie outside of the
spatial domain where convolutional and pooling structures can couple
image data.
From a grasping perspective, the left-middle and right-middle keypoints
are spatially separated but geometrically coupled.
The intent is for the non-local module to learn feature map dependencies
that contribute to joint keypoint detection and keypoint grouping.

Besides the keypoint heatmaps, this branch also generates embedding and
offset mappings.  
The offset mappings are the most relevant to keypoint detection, 
while the embedding is used later for grasp prediction
(see Section \subsecref{grouping}).
The keypoint heatmaps are output at a lower resolution than the input
images, which leads to quantized outputs when the keypoint coordinates
in the output domain are mapped to keypoint coordinates in the input
image domain.
The offset mapping corrects the quantization error due to
resolution mismatch between the input image dimensions and the heatmap
dimensions.
The correction is done by predicting
a small coordinate correction to the predicted pixel locations,
thereby
providing grasp keypoint coordinates with pixel-level resolution
in the input image coordinates, as obtained from heatmap maxima in lower
resolution coordinates.
Denote an offset mapping by $\omap:\Real^2 \rightarrow \Real^2$, as
encoded by the matrix $\omap \in \mathbb{R}^{2 \times H \times W}$.
There will be one for each \emph{grasp keypoint heatmap}, $\omap_L$ and
$\omap_R$, for left and right.

\subsubsection{The validation branch.}
The \emph{center keypoint heatmap} in the validation branch reflects
the probability distribution of grasp midpoints. The grasp midpoint is
agnostic to the orientation of the grasp, therefore there are only two
classes \emph{grasp} and \emph{no-grasp}. Given the opposing probability
relationship between these two classes, only a single heatmap is
required to capture the distribution (here, it is of the \emph{grasp} class). 
The \emph{center keypoint heatmap}
$\aux \hmap: \Real^2 \rightarrow \Real$ 
is stored as the matrix 
$\aux \hmap \in  \mathbb{R}^{H \times W}$.

\subsubsection{Network loss functions.} Using the same definition
for the variant of focal loss used in CornerNet \citep{law2018cornernet},
define the objective loss functions for the \emph{grasp keypoint heatmap}
and \emph{center keypoint heatmap} to be
\begin{align} 
  \eqlabel{loss_graspLR}
  \loss{det}{gk}(\hmat, \hmatGT)
    & = \frac{-1}{N}\sum_{c=1}^{C}\sum_{i=1}^{H}\sum_{j=1}^{W}
        \detCost(\hmat_{cij}, \hmatGT_{cij}; \alpha, \beta)\ \text{and} \\
  \eqlabel{loss_graspC}
  \loss{det}{cen}(\hmat, \hmatGT)
    & = \frac{-1}{N}\sum_{i=1}^{H}\sum_{j=1}^{W}
        \detCost(\hmat_{ij}, y_{ij}; \alpha, \beta),
\end{align}
where the heatmap detection loss function is
\begin{equation} \eqlabel{detCost}
  \hspace*{-0.5em}
  \detCost(\hmat, \hmatGT; \alpha, \beta) = 
    \begin{cases}
            (1-\hmat)^{\alpha} \log(\hmat) &  \text{if}\  y = 1\\
            (1-\hmatGT)^{\beta}\,\hmat^{\alpha} \log(1-\hmat) & \text{otherwise}
    \end{cases}
\end{equation}
for given hyperparameters $\alpha$ and $\beta$ that affect the
penalization of negative samples
(we set $\alpha$ and $\beta$ to be 2 and 4 for all experiments).
For the \emph{grasp keypoint heatmap}, the map 
$\gk{\hmap}:\oclass \times \Real^2 \rightarrow \Real$, indexed via
$\gk{\hmat}_{cij}$ for
the orientation class $c \in \oclass$ and the location $(i,j) \in \Real^2$,
describes the predicted heatmap score at $(i,j)$ for class $c$,
while $\gk{y}:\oclass \times \Real^2 \rightarrow \Real$ is the corresponding
ground truth. 
Since the \emph{center keypoint heatmap} has only one class, it is
$\aux{\hmap}:\Real^2 \rightarrow \Real$ with $(i,j) \in \Real^2$
indices specifying the evaluation coordinates. The same holds for the
ground truth center heatmap $\aux{\hmat}:\Real^2 \rightarrow \Real$.
The number $N$ in \eqref{loss_graspLR} and \eqref{loss_graspC} is the
quantity of grasp bounding boxes in the input image.

Assume that the reduction factor between the input image dimensions and
the heatmap dimensions is given by $\redFact$. Let $k$ index into the
ground truth keypoint for a given input image.  The pixel coordinates
$(i_k, j_k)$ of the keypoint determine the ground truth offset vector 
$\omatGT_{i_k j_k} \in \Real^2$ for the quantized heatmap coordinates,
\begin{align} \label{def_offset}
  \omatGT_{i_k j_k} = (\frac{i_{k}}{\redFact} - \left \lfloor \frac{i_{k}}{\redFact} \right \rfloor,
           \frac{j_{k}}{\redFact} - \left \lfloor \frac{j_{k}}{\redFact} \right \rfloor),
\end{align}
where $\lfloor \cdot \rfloor$ represents the floor operation.
The objective function for the offset map is the smooth $\smLone$ loss
\citep{ren2015faster}, written as $\smLone(\cdot)$, and leading to the
offset map loss functional
\begin{equation} \label{loss_offset}
  \loss{off}{gk} = \frac{1}{N} \sum_{k=1}^{N} 
                        \smLone(\omat_{i_k j_k}, \omatGT_{i_k j_k}),
\end{equation}
where $\omat_{i_k j_k}$ is the offset map output value at $(i_k, j_k)$
for a given input image and ground truth data containing $N$ keypoints.

\subsection{Keypoint Grouping \subseclabel{grouping}}

The detection step in the grasp hypothesis branch generates candidate
left-middle and right-middle keypoints. To generate grasp candidates,
these keypoints must be paired or grouped. Grouping will use the
embedding to confirm keypoint pairings. Conceptually, the embedding
\citep{newell2017associative} attempts to recover the association between
different grasp keypoints in a learnt embedding vector space.  When
combined with the orientation class predictions and the validation
branch, the keypoint grouping step produces ranked grasp candidates.

\subsubsection{The embedding.}
The associative embedding implementation used here was first implemented
for detecting human joints through an embedding vector for each joint 
\citep{newell2017associative}.  The distance between embeddings relates
to the likelihood that different joints are grouped together, and serves as
a similarity score.  Here, the same process is repurposed to generate
embeddings for the left-middle and right-middle keypoints with the aim
of outputting similar values for paired keypoints and dissimilar values
for unpaired keypoints.  As a function, the embedding maps the network
feature space to a scalar embedding value, 
$\emap:\featSpace \rightarrow \Real$, where $\mathcal{F}$ is the network
feature space co-domain of the non-local module.
The processing generates multiple left-middle and right-middle grasp keypoints
associated with the 1D embedding values, each of which has a left-middle
and right-middle embedding, denoted $\emap_L$ and $\emap_R$. 
%

\subsubsection{The embedding loss.}
Learning the embedding involves promoting keypoints belonging to the
same grasp to output similar scalar values, while forcing ones belonging
to different grasps to output dissimilar values.
Since the embedding attempts to connect the spatially separated
left-middle and right-middle grasps, the non-local module described
earlier can contribute to this coupling by inducing dependencies across
all pixels in the feature map.
%
Using competing \emph{pull} and \emph{push} losses achieves
similarity/dissimilarity learning, such that backpropogating errors
through the non-local module induces the necessary dependencies.
The \emph{pull} loss drives the network to predict similar embeddings
for keypoints belonging to the same grasp bounding boxes. 
The \emph{push} loss drives the network to separate keypoints 
belonging to different grasp bounding boxes,
Each grasp $k$ contained in an image from the ground truth dataset has
associated to it two hypothesized embedding values $\emat_{L}^k$ and
$\emat_R^k$ based on the network's feature vector output $f_L^k, f_R^k
\in \featSpace$ at that keypoint's location. 
The \emph{pull} and \emph{push} losses are

\begin{align} \label{loss_pull}
  \loss{pull}{gk} & = \frac{1}{N} \sum_{k=1}^{N} [(\emat_{lm_{k}}-\emat_{k})^2 +
  (\emat_{rm_{k}}-\emat_{k})^2]\\
  \label{loss_push}
   \loss{push}{gk} & = \frac{1}{N(N-1)} \sum_{k=1}^{N} \sum_{j=1,j\neq k}^{N} 
   \text{max}(0, 1 - \left | \emat_{k} - \emat_{j} \right |)
\end{align}
where $\emat_k$ is the mean value of $\emat_{lm_{k}}$ and $\emat_{rm_{k}}$
and $\emat_j$ is the mean value of other keypoint pairs.

\subsubsection{Grasps as paired keypoints.}
When multiple keypoint prediction classes have geometric constraints
between them, keypoint estimation benefits from defining an anchoring
keypoint, usually defined to be the center or some other center-like
location \citep{duan2019centernet}. The \emph{center keypoint heatmap}
plays that role here.  Each pixel on the \emph{center keypoint 
heatmap} presents the probability of being a center point for some 
left-right keypoint pair.
Initially grasp keypoint pairs are determined purely by the grasp
orientation class and embedding values. Pairs with large differences
between the embedding values of the two keypoints will be removed (i.e.,
less than $\threshEmbed$). The center points from the candidate
left-middle and right-middle keypoint pairs provide a follow-up culling
step by obtaining the confidence scores of the center coordinate from
the \emph{center keypoint heatmap}. Pairs leading to a low confidence
are removed (i.e., less than $\threshCenter$).

\begin{algorithm}[t]
\textbf{Input:} \\
Two \emph{grasp keypoint heatmaps} $\gk \hmap_L$ and $\gk \hmap_R$ with associated embedding heatmaps $\emap_L$ and $\emap_R$, and offset heatmaps $\omap_L$ and $\omap_R$;\\
A \emph{center keypoint heatmap} $\aux \hmap$ \\
\textbf{Output:} \\
A set of 4-DOF grasp representations $G$ 
  \begin{algorithmic}[1]
    \State $G^{gk}_{lm} = Select\_Grasp\_Keypoint(\gk \hmap_L, \emap_L, \omap_L)$
    \State $G^{gk}_{rm} = Select\_Grasp\_Keypoint(\gk \hmap_R, \emap_R, \omap_R)$
    \State $G_{pair} = Extract\_Center\_Keypoint(\aux \hmap, G^{kp}_{lm}, G^{kp}_{rm})$
    \State $G_{candidate} = Post\_Process(G_{pair})$
    \State $G = Orient\_Filter(G_{candidate})$
  \end{algorithmic}
\caption{Overall algorithmic processing flow for keypoint grouping.
  \label{algGrouping}}
\end{algorithm}

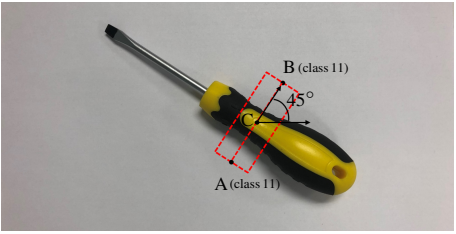
\begin{figure}[t]
    \centering
    \begin{tikzpicture} [outer sep=0pt, inner sep=0pt]
    \node[anchor=north west] (a) at (0in,0in) 
    {\includegraphics[width=0.7\columnwidth,clip=true,trim=0in 1.0in 0in 0.8in]{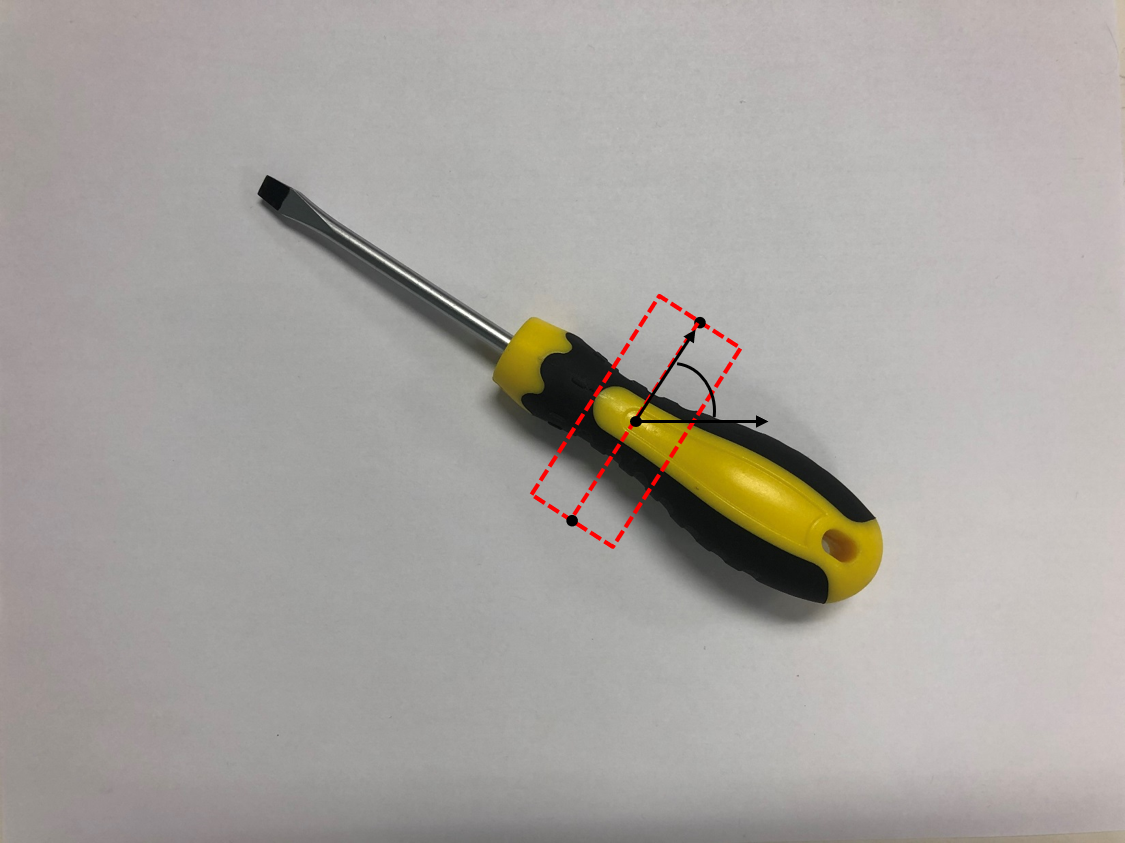}};
    \node[text width=2.0cm] at (3.8, -2.4) {\scriptsize A};
    \node[text width=2.0cm] at (4.0, -2.42) {\tiny (class};
    \node[text width=2.0cm] at (4.43, -2.42) {\tiny 11)};
    \node[text width=2.0cm] at (4.7, -0.85) {\scriptsize B};
    \node[text width=2.0cm] at (4.9, -0.87) {\tiny (class};
    \node[text width=2.0cm] at (5.33, -0.87) {\tiny 11)};
    \node[text width=2.0cm] at (4.15, -1.55) {\scriptsize C};
    \node[text width=2.0cm] at (4.75, -1.29) {\scriptsize 45};
    \node[text width=2.0cm] at (4.99, -1.28) {\scriptsize $\degree$};
    \end{tikzpicture}
    \caption{Orientation filtering intends to filter out grasp pairs whose difference between continuous and discrete orientation is over a certain threshold.
    \label{fig_ori_filter}}
\end{figure}

\subsection{Orientation Filter \label{secOrFilter}}
By quantizing the continuous orientation into discontinuous classes, 
the multi-class classification of grasp orientation might suffer from
discontinuities. Furthermore, performing dense pixel-wise orientation
classification might lead to errors through missing global visual
information.  To address potential issues from the discrete orientations
and regional information, a continuous grasp orientation hypothesis is
directly computed via grouped keypoints and provides complementary
information for comparison.
The keypoint grouping process outputs a set of grasp candidates, each
represented by the tuple 
$\pregrasp = (\point_L, \point_R, \orient, \gconf)$ 
composed of the left and right grasp keypoints, the shared orientation
class, and the center keypoint confidence score, respectively.
At this step, there are two distinct forms of orientation information
contained in each grasp candidate tuple. 
One is the discrete orientation $\theta_{1}$ from class $\oel \in
\oclass$ and the second is the continuous orientation $\theta_2$ based
on the left and right keypoints, 
\begin{align} 
  \label{discrete_orientation}
  \theta_1 & = \frac{\pi}{\left| \oclass \right|} \orient - \frac{\pi}{2}
    & \in {[-\frac \pi 2, \frac \pi 2]}, \\
  \label{continuous_orientation}
  \theta_2 & = \angle(\point_R - \point_L)
    & \in {[-\frac \pi 2, \frac \pi 2]}.
\end{align}
The orientation filtering step eliminates grasp candidates whose
absoluate difference between the discrete and continuous orientations is
larger than a threshold, $\threshOrient$. 
%
For the candidate grasp shown in Fig. \ref{fig_ori_filter}, the
classification of two grasp keypoints is 11 whereas its continuous
orientation is 45\degree. Presuming that the class label 11 corresponded
to an orientation of 30\degree, then the detection would be discarded
due to the large difference between the two orientations.  If instead
11 corresponded to 45\degree, then the grasp would be kept.

\subsection{The GKNet Procedure\subseclabel{GKNet}}

This section describes the Grasp Keypoint Network procedure for
collecting keypoint candidates from visual input and then grouping them
into grasp candidates, with the steps performed after the network
output summarized in Algorithm \refsp{algGrouping}.  
The grouping process takes as input the \emph{grasp keypoint}, embedding, 
and offset heatmaps from the \emph{grasp hypotheses branch}, and the
\emph{center keypoint heatmap} from the \emph{validation} branch, 
then provides a set of grasp representations (\ref{eqn_grasp_representation}). 
Step 1 selects the top-$\topk$ keypoints over the predicted left \emph{grasp
keypoint heatmap} based on the confidence score and extracts the
associated embedding values and offsets at the locations of selected
keypoints  
($k = 100$ in all experiments).
This step leads to a set of keypoint descriptions given by
$\gk G_{\rm lm} = {(x_{\rm lm}, y_{\rm lm}, o_{x}, o_{y}, c, s, v_{embed})}$ 
where 
$x_{lm}$ and $y_{lm}$ are the keypoint coordinates, 
$o_{x}$ and $o_{y}$ are the coordinate offsets, 
$c$ is the quantized orientation class, 
$s$ is the confidence score and 
$v_{embed}$ is the embedding value. 
Step 2 applies the same procedure for the right \emph{grasp keypoint heatmap}. 
In Step 3, the selected left and right keypoints combine to create
$\topk^{2}$ pairs. Each pair has a confidence score obtained from the
\emph{center keypoint heatmap} based on the computed location of
the center point between the left and right coordinates. 
Step 4 filters the $\topk^{2}$ grasp pairs into coarse grasp candidates
based on three conditions 
(1) the orientation classes of the left and right keypoints should agree, 
(2) the difference between embedding values of the left and right
keypoints should be less than a given threshold $\threshEmbed$, 
and 
(3) confidence scores of center points should be above a threshold $\threshCenter$. 
In Step 5, the remaining grasp candidates go through the orientation filter
algorithm with passing candidates forming the set of hypothesized good
grasps. 
The number of output final grasps is set to be 100 in all experiments.

\section{Vision Benchmarks and Evaluation \label{sec:VB}}
The Grasp Keypoint Network approach to grasp recognition is a visual
processing approach intended to integrate with a robotic
manipulation pipeline. Consequently, evaluation of the approach will
involve both benchmarking as a visual input/output method, and as a
component of a manipulation system with an actual manipulator.
Visual processing will consider the accuracy of the output and the
timing of the method relative to other published one-stage, two-stage,
grasp quality and other deep learning approaches \citep{jiang2011efficient, 
asif2017rgb, lenz2015deep, mahler2017dex, morrison2019learning, wang2016robot, 
redmon2015real, kumra2017robotic, guo2017hybrid, zhou2018fully, chu2018real, 
asif2019densely}.

Prior to discussing the outcomes of the various experiments, this
section describes the GKNet training process, the experimental
evaluation criteria, and any additional common methodological elements
associated to the vision-only experiments. Experimental setup and evaluation
metric for physical experiments will be discussed in Section \ref{sec:PE}. 

\subsection{Datasets and Preprocessing \label{secDataTrain}}

There are two benchmark datasets commonly used for training and testing
grasp detection networks: the Cornell and Jacquard datasets 
\citep{cornell2013,depierre2018jacquard}.
The Cornell dataset contains 885 RGB-D images of 244 different objects,
with multiple images taken of each object from different poses. The
image dataset was collected and manually annotated with 5,110 positive
and 2,909 negative grasps. Though the Cornell dataset is used as a
standard dataset, annotation is relatively sparse and its size
is small. 
Based on a subset of ShapeNet \citep{chang2015shapenet}, the
simulation-based Jacquard dataset has more than 50,000 images of
11,000 objects, with 1 million {\em positive} grasps.  

\subsubsection{Cornell Dataset.}
Following other RGB-D implementations already trained on pre-existing
backbones for a different task (e.g., object
detection), we pass to the network the RG-D channels only by using the
depth channel as the blue channel input \citep{redmon2015real}. 
The depth information is normalized to lie in the same range, $[0,255]$,
as the RGB data.  All input channels are re-scaled to lie in the range
$[0,1]$, followed by a whitening process (per channel, per pixel, mean
value subtraction with division by the standard deviation).  The
obtained mean values for the RG-D channels are (0.85, 0.81, 0.25), with
standard deviations of (0.10, 0.11, 0.09).
The image resolution and annotation coordinates are remapped to the
matrix dimensions 256$\times$256, followed by application of the data
augmentation strategy presented in \citep{chu2018real} to increase the
amount of annotated input/output data and avoid model overfitting.

\begin{figure}[t]
  \centering
  \begin{tikzpicture} [outer sep=0pt, inner sep=0pt]
  \scope[nodes={inner sep=0,outer sep=0}] 
  \node[anchor=north west] (a) at (0in,0in) 
    {\includegraphics[width=3cm,clip=true,trim=0in 0in 0in 0in]{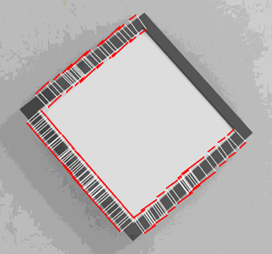}}; 
  \node[anchor=north west, yshift=-0.1cm] (b) at (a.south west) 
    {\includegraphics[width=3cm,clip=true,trim=0in 0in 0in 0in]{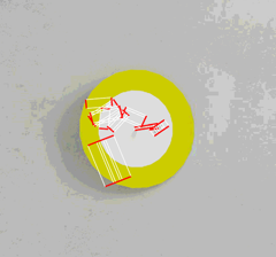}}; 
  \node[anchor=north west, xshift=0.1cm] (c) at (a.north east) 
    {\includegraphics[width=4.53cm,clip=true,trim=0in 0in 0in 0in]{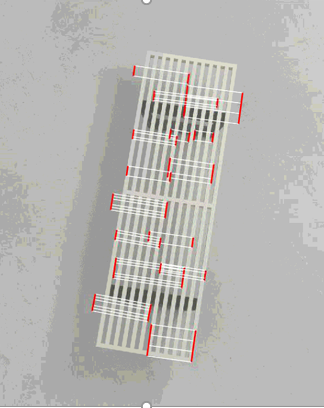}};
  \node[anchor=north west, yshift=-0.1cm] (d) at (b.south west) 
    {\includegraphics[width=7.63cm,clip=true,trim=0in 0in 0in 0in]{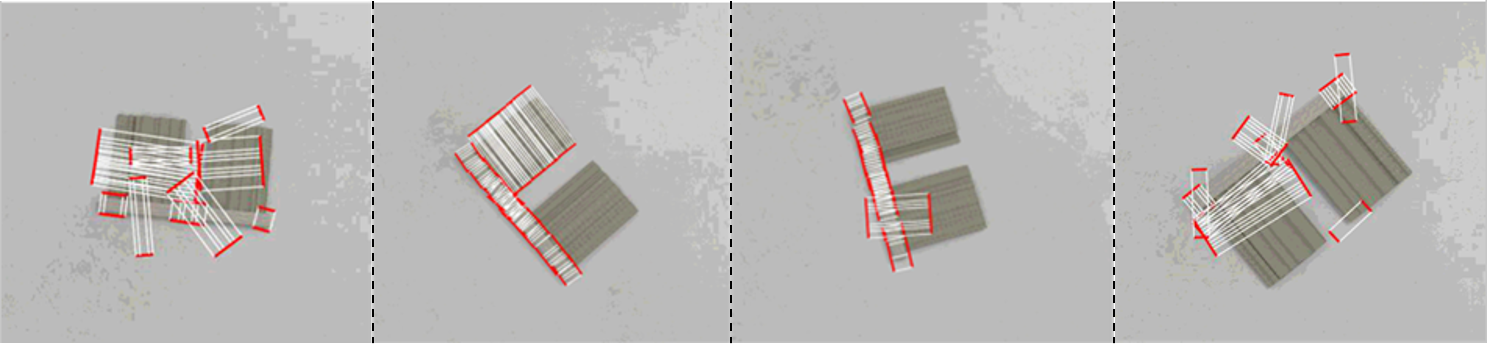}}; 
  \endscope
  \foreach \n in {a,b,c,d} { 
    \node[anchor=north west,xshift=2pt,yshift=-2pt] at (\n.north west) {(\n)};
  }
  \end{tikzpicture}
  \caption{
  (a)-(d): Four issues with the Jacquard dataset.
    \label{fig_jac_fail_case}}
  \vspace*{-1.0ex}
\end{figure}

\subsubsection{Jacquard Dataset.}
The Jacquard dataset was produced through automatically generated
annotations in order to have a much larger training set relative to the
Cornell dataset without the manual annotation demands.  Done in
simulation, the grasps were obtained via a sample and confirm process.
Compared with the Cornell dataset, images in the Jacquard dataset are
annotated with much denser grasping bounding boxes, however they are not
all physically meaningful. 
While the Cornell dataset has many missing (false negative) annotations,
the Jacquard dataset has several false positive annotations as well as
missing annotations.
After reviewing the dataset, we identified the following four problems
with the annotations:
(1) They can be dense on one side but sparse or even missing on the other
  side for some objects with symmetric shapes, 
  see Fig.  \sfigref{fig_jac_fail_case}a; 
(2) Some do not seem plausible and may be a function of incorrect
  physics within the simulator, see Fig. \sfigref{fig_jac_fail_case}b; 
(3) The sampling process can lead to sparse annotations for certain
  objects with multiple, continuous grasp regions, see
  Fig. \sfigref{fig_jac_fail_case}c, 
(4) Orientation dependent grasp annotations lead to highly variable
grasp annotation outcomes for the same object, Fig.
\sfigref{fig_jac_fail_case}d. 
These issues negatively influence the training process and the test evaluation
processes. 

To correct for these issues, we defined a filtering strategy for
selecting a subset from the Jacquard dataset.
The filter strategy follows three rules based on the ratio of the union
of all grasp regions to the masked region of the object: 
(a) Keep annotation data whose ratio is larger than 0.8;
(b) Remove annotation data whose ratio is less than 0.2; and 
(c) Manually check images whose ratio is between 0.2 and 0.8 and remove
  those images whose annotations satisfy either one of the above four
  cases.
We call the resulting selected subset the Abridged Jacquard Dataset
(AJD), which is publicly available \citep{AJD}. The AJD consists of 
46{,}857 images of 10{,}799 objects, with 4{,}420{,}953 positive 
grasp bounding boxes in total.

Due to downsampling of output keypoint heatmaps relative to the input
image, multiple grasp keypoints from different grasp instances can map
to the same pixel in the ground truth heatmap.
For example, if the left-middle point of grasp instance $A$ is within
the radius of downsampling ratio of the left-middle point of grasp
instance $B$ on the input image, they will be located at the same pixel
on the groundtruth heatmap. 
To deal with the resulting ambiguity, we refine the dense annotations 
by always selecting the first grasp bounding box processed
to make sure that there will not be any points from different grasp
bounding boxes mapping to the same pixel on the downsampled map.
The above strategy isn't applied to the Cornell dataset, since its
grasp annotations are sparse. 

Image normalization follows the same procedure applied to the Cornell
dataset. The mean values of the RG-D channels are (0.71, 0.71, 0.20),
and the standard deviations are (0.06, 0.07, 0.09).  The Jacquard images
and annotations were remapped to the matrix dimensions 512$\times$512,
with no data augmentation applied. 

\subsection{Deep Network Architecture and Training}

We use Deep Layer Aggregation (DLA) as our backbone network to extract
features from input images \citep{yu2018deep}.  
DLA augments standard
neural network architectures with an iterative and hierarchical aggregation strategy to fuse semantic and spatial information across layers. 
The deep layer aggregation structures helps network capture deeper and wider information for recognition and localization. 
Meanwhile the hierarchy aggregation architecture requires fewer parameters, which reduces size of the entire network and inference time. 
We choose DLA as our final backbone network since it achieves the best
performance. 
Rich semantic and spatial information improves performance
of keypoint detection and fewer network parameters enables GKNet to run
in real-time. 
However there is no restriction on the backbone network
for GKNet; it can be switched to other state-of-the-art keypoint
detection neural networks.

Our work is implemented in PyTorch 1.1.0 \citep{paszke2017automatic} and the 
backbone network DLA is pretrained on ImageNet \citep{deng2009imagenet}. 
We use random affine transformation as data augmentation and use Adam 
\citep{diederik2015adam} to optimize training loss:
\begin{align} \label{training_loss}
  \loss{}{} = \loss{det}{gk} + \loss{det}{cen} 
                + \alpha \loss{pull}{gk} + \beta \loss{push}{gk} 
                + \gamma \loss{off}{gk},
\end{align}
where $\alpha$, $\beta$ and $\gamma$ are the weights of the pull, push
and offset loss for \emph{grasp keypoint heatmap}. 
For training, we use a batch size of 8 and train our ntwork end-to-end
on a single Titan XP. 
Training a network model with the Cornell dataset runs for 20 epochs
with an initial learning rate of $1.25 \times 10^{-4}$ that gets divided
by 10 at the 10th and the 15th epoch. 
Training a network model with the AJD runs for 30 epochs with the same
initial learning rate and with divisions at the 20th and the 25th epoch.

\subsection{Parameters}

We report some essential hyper-parameters used in training and testing
process here; the rest are available in the released code \citep{ivagit_GKNet}. 
With a downsampling ratio of $\redFact = 4$, GKNet outputs heatmaps with
size of 57$\times$57 and 128$\times$128, for the Cornell and Jacquard
dataset respectively. 
For all training processes, we choose weighting coefficients in
(\ref{training_loss}) to be $\alpha = 1$, $\beta = 1$ and $\gamma = 1$.
Due to the different image source and grasp annotation distributions,
we set different testing stage filtering coefficients 
$\threshEmbed$, $\threshCenter$ and $\threshOrient$ for the 
datasets.

\vspace*{0.25em}
\noindent
\textbf{Cornell.}
We set 
$\threshEmbed = 1.0$, 
$\threshCenter = 0.05$ and 
$\threshOrient = 0.24$. 
The number of orientation classes is $\left| \oclass \right| = 18$.

\vspace*{0.25em}
\noindent
\textbf{AJD.}
We set 
$\threshEmbed=0.65$, 
$\threshCenter=0.15$, and 
$\threshOrient=0.1745$. 
The number of orientation classes is $\left| \oclass \right| = 36$.

\subsection{Evaluation Metrics for Vision Benchmarks}
The standard grasp evaluation metric considers the full grasp rectangle 
\citep{lenz2015deep}.  By design, the GKNet grasp representation
lacks the variable $h$ describing the bounding box. To apply the same
evaluation metric as other publications, the value $h$ is set to the
average of all grasp candidates for the training dataset used.  These
values work out to be 23.33 and 20 for the Cornell dataset and the AJD,
respectively.  The predicted grasp candidate is a positive match if:
\begin{enumerate}
  \item the orientation difference between predicted grasp and
  corresponding ground truth is within 30$^\circ$; and 
  \item the Jaccard index of the predicted grasp $g_p$ and the ground truth grasp $g_t$ is greater than 0.25:
    \begin{equation} \label{eqn_jaccard_index}
      J(g_p,g_t) = \frac{\vert g_p \cap g_t \vert}{\vert g_p \cup g_t \vert} 
                 > \frac 14.
    \end{equation}
\end{enumerate}
The speed of the grasp recognition process is measured in frames per
second (fps).

\begin{figure*}[t]
  \centering
  \begin{tikzpicture} [outer sep=0pt, inner sep=0pt]
  \scope[nodes={inner sep=0,outer sep=0}] 
  \node[anchor=south east] (a) 
    {\includegraphics[width=4.25cm,clip=true,trim=0.9in 1.25in 1.25in 1.0in]{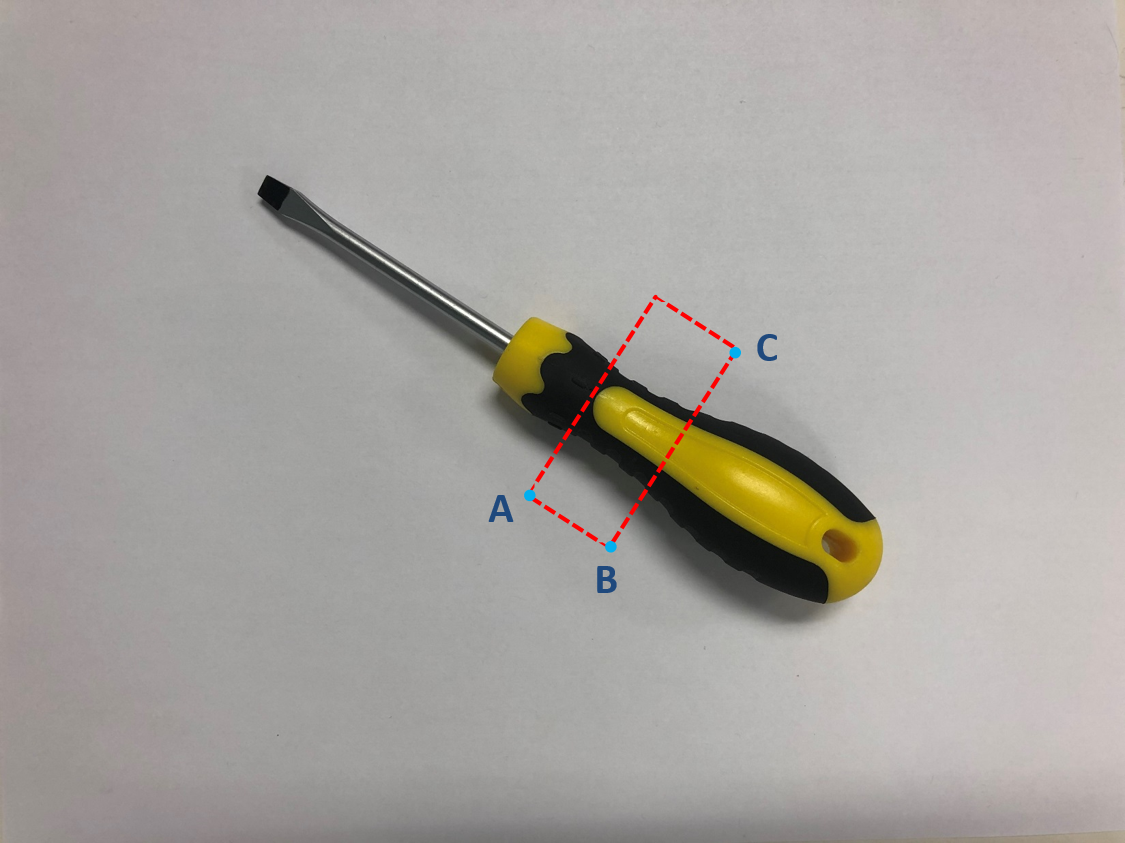}}; 
  \node[anchor=south west, xshift=0.1cm] (b) 
    {\includegraphics[width=4.25cm,clip=true,trim=0.9in 1.25in 1.25in 1.0in]{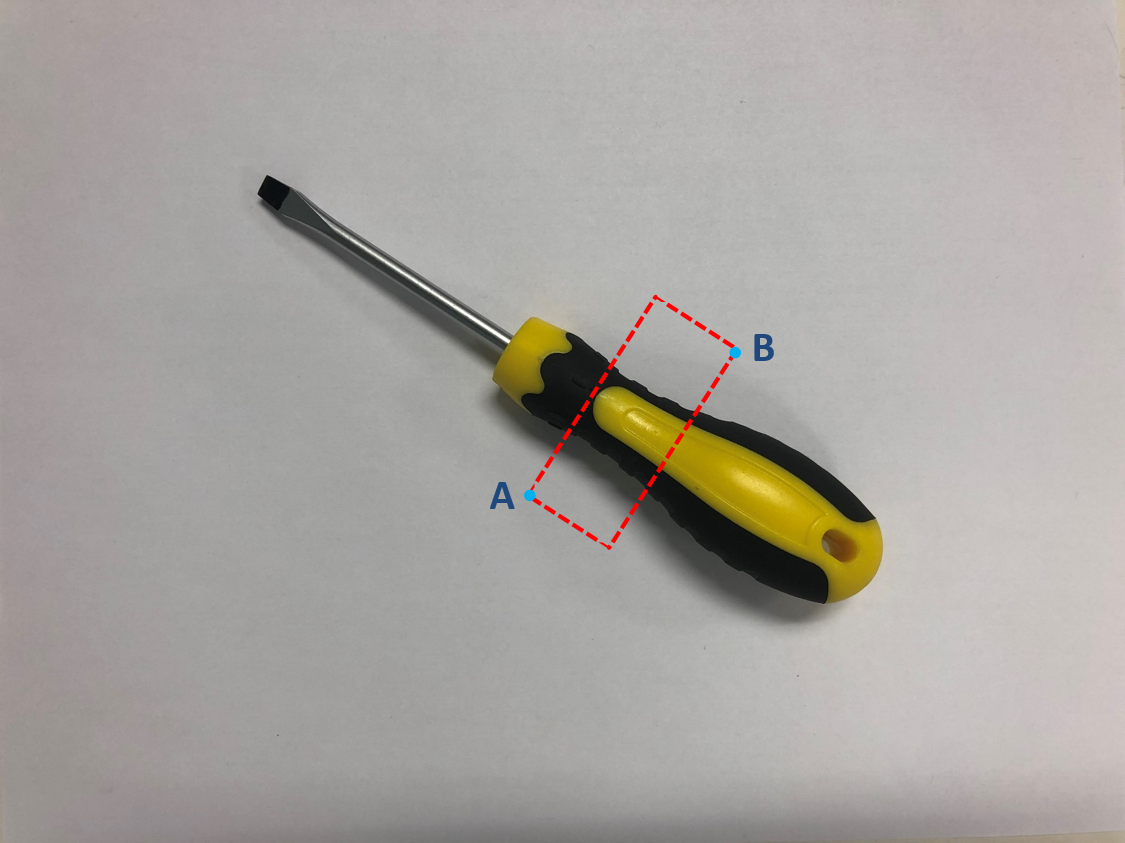}}; 
  \node[anchor=north west, xshift=0.1cm] (c) at (b.north east)
    {\includegraphics[width=4.25cm,clip=true,trim=0.9in 1.25in 1.25in 1.0in]{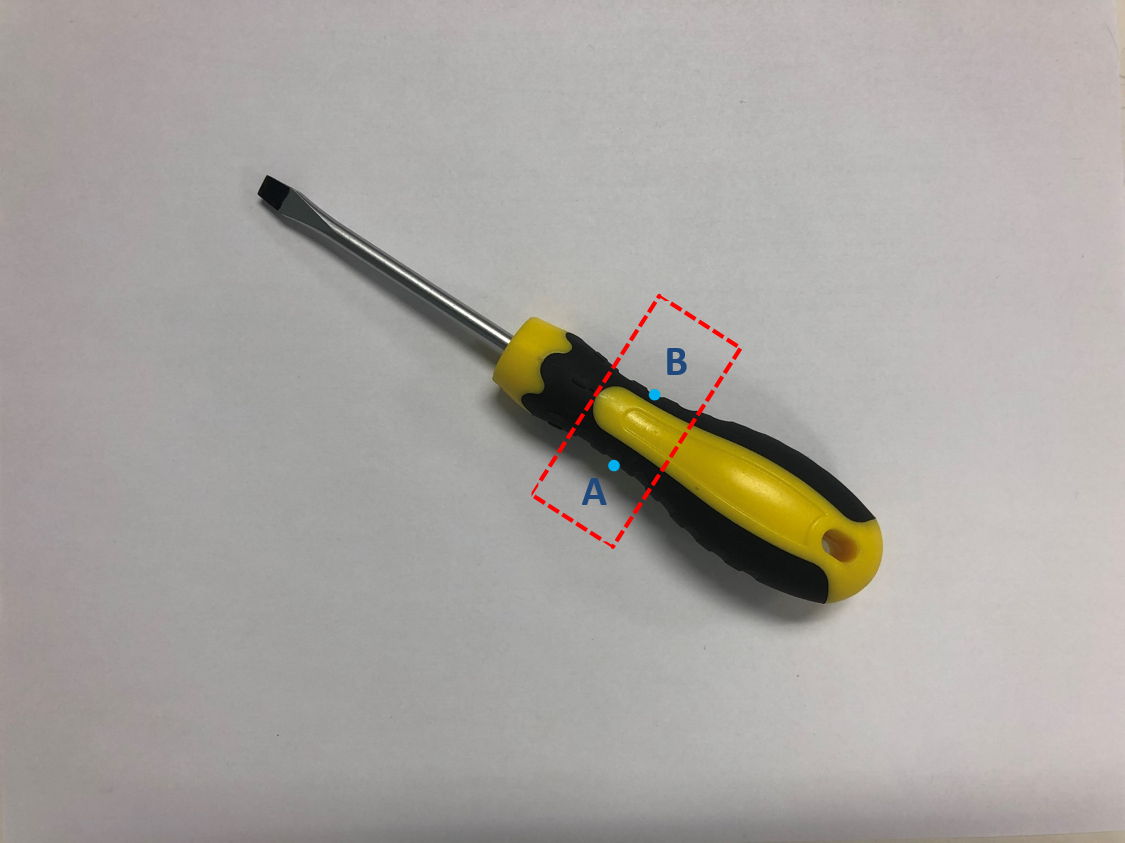}};
  \node[anchor=north west, xshift=0.1cm] (d) at (c.north east)
    {\includegraphics[width=4.25cm,clip=true,trim=0.9in 1.25in 1.25in 1.0in]{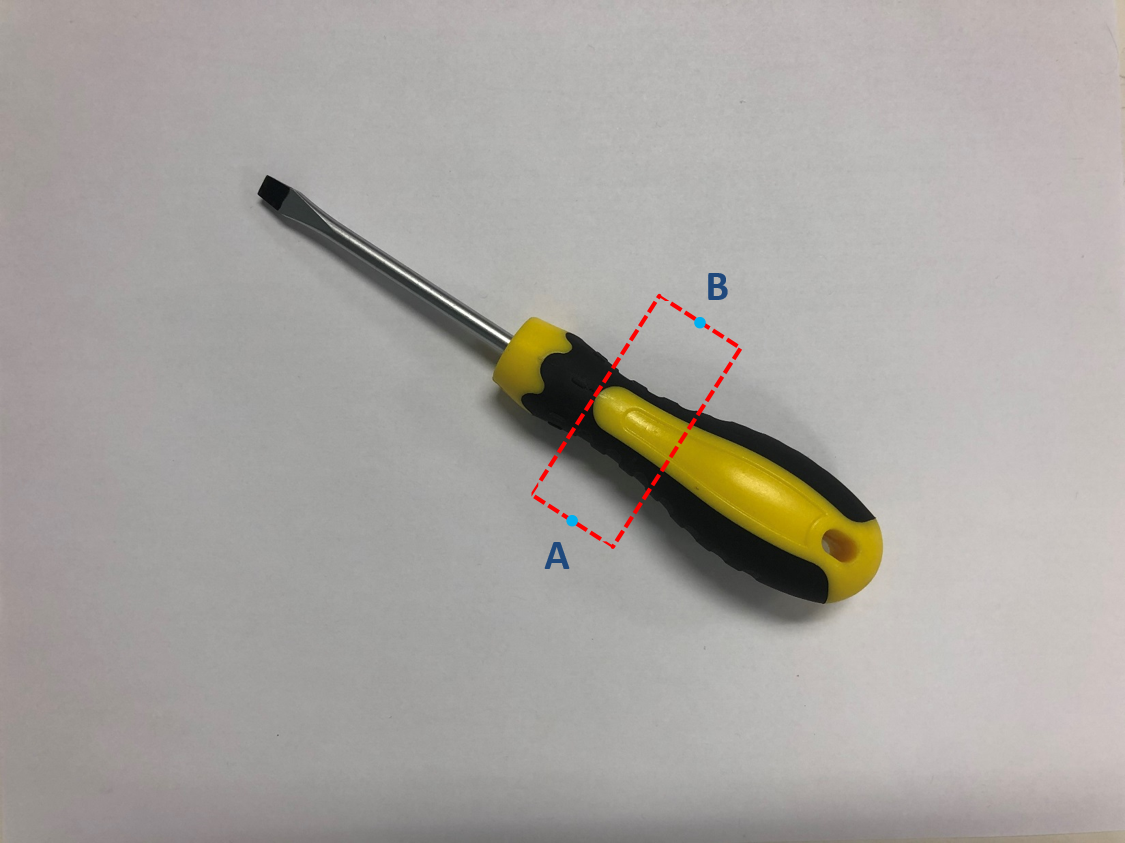}}; 
  \endscope
  \foreach \n in {a,b,c,d} { 
    \node[anchor=north west,fill=white,draw=white,inner sep=2pt] at (\n.north west) {(\n)};
  }
  \end{tikzpicture}
  \caption{
  (a)-(d): Grasp bounding box representations with 
    three corner points, 
    two corner points, 
    two edge points, and 
    two middle points.  Benchmarking on the AJD indicates that using the
    two middle points provides the best outcomes.
    \label{fig_4_grasp_reps}}
  \vspace*{-1.0ex}
\end{figure*}

\section{Vision Benchmark Results \label{sec:Vision}}

Assessment of GKNet as a visual processing algorithm tests two
aspects: the network design choices and the final design for grasp
prediction.  Regarding design elements, there is the assertion that the two
keypoint representation with left-middle and right-middle keypoints is a
preferred representation over alternative versions (\S
\ref{secAblationKP}). Secondly, different backbone options are possible
with similar properties (\S \ref{secBBone}).
The two studies will lead to the final chosen design for GKNet.
Benchmark evaluations will explore the hypothesis that a keypoint-based,
one-stage neural network structure can provide the best outcome relative
to performance and processing speed. GKNet is compared to other
published works using the Cornell dataset (\S \ref{secGraspOutCD}), and
with publicly available methods using the AJD (\S \ref{secGraspOutAJD}).
Visualizing published methods on a performance versus processing frame
rate graph shows that GKNet is closer to the extremal corner (high
performance and high frame rate) than other methods, thereby confirming
the assertion of balancing accuracy and speed.

\subsection{Grasp Keypoint Representation Study \label{secAblationKP}}
There is some flexibility when choosing the keypoint representation for
recovering grasp bounding boxes, four of which are depicted in
Fig.~\ref{fig_4_grasp_reps}. The first two preserve the full grasp
representation coordinates, while the last two provide less information
related to the lengths of the bounding box edges.
To confirm that the chosen middle points representation provides the
best outcomes, grasp keypoint networks with the other representations
were also created and trained using both the Cornell dataset and AJD,
individually.  Since the Cornell dataset does not provide segmentation
masks, the two edge points representation,
Fig.~\ref{fig_4_grasp_reps}(c), cannot be trained due to missing
ground-truth information.  Hence, the missing entry for LeRe in Table
\tabref{eodgr} for outcomes using the Cornell dataset.

There are 4 different grasp representations and one more variation based
on the grasp representation in Fig.~\ref{fig_4_grasp_reps}(b), which
yields 5 outcomes for each dataset in the Table \tabref{eodgr}.
When using three corner points, Fig.~\ref{fig_4_grasp_reps}(a), the
grouping process will be more complex since additional constraints or
factors must be included to properly group the point triplet from the
heatmap outputs. They introduce an additional process that can add false
positive and negative outcomes.
For the top-left and bottom-right corner points,
Fig.~\ref{fig_4_grasp_reps}(b), there is no unique bounding box due to
the unknown slope of the sides. Here, the estimated slope or orientation
of the bounding box edges must be inferred from the discrete orientation
classification information in the output grasp keypoint heatmaps.  
Considering that employing classification to recover orientation
inevitably introduces quantization error, we explore a variation to
address the quantization error, which incorporates one more branch to
predict another heatmap for the orientation offset. The final offset
for grouped grasp candidate will be computed by averaging the
offset predicted for each grasp keypoint.
The third representation with left-edge and right-edge points,
Fig.~\ref{fig_4_grasp_reps}(c), was explored as an option based on the
premise that object edge points and their neighboring regions contain
potentially useful visual information, which could help the network
learn visual features of grasp keypoints and support generalization. 
It contains partial information regarding the bounding box edge lengths
(a minimum value for one edge and no information for the other).
The fourth representation is the one chosen for GKNet. 

\begin {table}[t]
  \centering
  \vspace*{1.5ex}
  \caption {Evaluation of different grasp representations on the Cornell and AJD. 
    TlBlBr: top-left, bottom-left and bottom-right keypoints; 
    TlBr: top-left and bottom-right keypoints; 
    TlBrOri: top-left and bottom-right keypoints with orientation offset; 
    LeRe: left-edge and right-edge keypoints; and 
    LmRm: left-middle and right-middle keypoints.
  \tablabel{eodgr}}
  \small
  \setlength{\tabcolsep}{4.5pt}
  \begin{tabular}{ | c | c | c | c | c | c |}
    \hline
    {\bf Rep.}  & {\bf TlBlBr}  & {\bf TlBr}  & {\bf TlBrOri}  & {\bf LeRe}  & {\bf LmRm}          \\ \hline
                       & \multicolumn{5}{c|}{Prediction Accuracy (\%)}        \\ \hline
    {\bf Cornell}         &  94.50  & 92.41 &  85.29   & N/A   & 96.90         \\ \hline
    {\bf AJD}             &  95.89  & 96.38 &  95.34   & 97.02 & 98.39         \\ \hline  
  \end{tabular}
\end {table}

All outcomes are recorded in Table~\tabref{eodgr}\,. 
The best performing grasp representation on the Cornell dataset is the
LmRm representation.  The same holds for the AJD.
For the other representations, the Cornell and AJD have opposing trends.
For the Cornell dataset, when excluding the LmRm options, performance
drops from left to right. For AJD it roughly increases if the TlBrOri
option is ignored. 
The discrepancy may be a result of the Cornell dataset having sparser
annotations than the AJD.
The drop for TlBrOri relative to TlBr for both
datasets suggests that this orientation ambiguous solution cannot be
enhanced by additional orientation information. 
The performance drop shows the difficulty of learning to directly
regress small orientation offsets in these circumstances; paired
corner representations and angle quantizations should be avoided.

Focusing on the AJD results, one inference is that keypoint grouping
plays an important role in the grasp keypoint detection. Performance
lowers with more keypoints, which has the added difficulty of correctly
grouping detected keypoints. The performance boost between the 
TLBr and LeRe representations for AJD shows implicitly encoded rotation
information contributes to grasp bounding box estimation and that the
quantization error introduced by classification degrades grasp
detection. 

The LmRm representation 
has the smallest set of geometric constraints to impose and the most
amount of orientation information contained in the network outputs. Both
properties contribute to the improved performance. The simplified
representation is possible because the edge length variable $h$ was
removed from the grasp coordinate vector. 

\subsection{Ablation Studies \label{secBBone}}
Three ablation studies performed are described here. 
First, several deep neural network backbones are selected to replace the
feature extractor of the proposed network.  
The goal of this test is to evaluate the
effectiveness and robustness of our proposed keypoint-based grasp
representation across backbone network architectures.
Second, for the purpose of evaluating the orientation filter, each
trained model will be tested with and without the proposed orientation
filter.
Third, to evaluate the effectiveness of lower resolution heatmaps with
offset mappings, the proposed method will be trained with a variant
that predict grasp keypoint heatmaps of the same dimenions as the
input image.

\textbf{Backbone Study.}
Various backbone options are possible. The performance impact should not
vary much since many networks have good learning properties.  However,
the execution speed and training properties will differ based on the
complexity of the network. 
Five network types were tested,
Deep Layer Aggregations (DLA) \citep{yu2018deep} (size 34),
Hourglass \citep{newell2016stacked} with two sizes (52 and 104),
ResNet \citep{he2016deep} with two sizes (18 and 50), 
VGG-16 \citep{simonyan2015very}, and
AlexNet.
AlexNet, VGG-16 and ResNet are chosen for comparison due to their
contributions to the evolution of convolutional neural networks.
AlexNet uses Rectified Linear Units (ReLu) in place of the tanh 
function.
VGG-16 use 3x3 filters to replace the large convolution filters in the
intial layers and had increased depth.
ResNet use a simple skip connections strategy and has fewer 
layers relative to other contemporary networks. 
Hourglass-52 and Hourglass-104 (Hg-52 and Hg-104) are chosen to explore
any potential performance boosts due to a deeper and more connected
network, at the cost of inference speed. 
There are two depth types chosen for ResNet and Hourglass in order to
explore the effect of depth for the same network architecture.
All models were tested with two variants: with and without the orientation
filter component (\S \ref{secOrFilter}).

\begin {table}[t]
  \centering
  \small
  \caption {GKNet Backbone: Cornell Dataset \tablabel{bbnets} }
  \addtolength{\tabcolsep}{-2pt}
  \begin{tabular}{ | l | c | c | c |}
    \hline
    \textbf{Approach} & \textbf{Acc. w o.f.} (\%) & \textbf{Acc. w/o o.f.} (\%) & \textbf{Speed} (fps) 
    \\ \hline
    Alexnet      &    95.0       &     94.8      & 83.33          \\ \hline
    VGG-16       &    96.8       &     96.4      & 55.56          \\ \hline
    Resnet-18    &    96.0       &     95.7      & 66.67          \\ \hline
    Resnet-50    &    96.5       &     96.4      & 52.63          \\ \hline
    Hg-52        &    94.5       &     93.6      & 33.33          \\ \hline
    Hg-104       &    95.5       &     95.3      & 21.27          \\ \hline
    \textbf{DLA} & \textbf{96.9} & \textbf{96.8} & \textbf{41.67} \\ \hline        
  \end{tabular}
\end {table}

Table \tabref{bbnets} contains the outcomes for the Cornell dataset
(image-wise split) and Table \tabref{bbajd} contains the outcomes for
the Abridged Jacquard dataset. 
All models are ordered based on their release date.
First, they show that the DLA backbone is both accurate and efficient.
Runtimes close to or surpassing the typical camera frame-rate
($\sim$20Hz) are possible. 
Second, among AlexNet, VGG-16 and ResNet, AlexNet achieves the worst
performance.
Models of VGG-16 and ResNet achieve similar performance. 
This result shows that skip connection might not lead to significant
improvement in keypoint detection.
Third, the improvements from ResNet-18 to ResNet-50 and Hourglass-52 to
Hourglass-104 show that deeper network architectures help capture rich
image features and improve the detection accuracy. However, the speed
loss may outweight the performance boost.
Fourth, Hourglass models do not achieve strong performance
compared to other networks with relatively simpler designs.
The Hourglass network is composed of a large amount of skip connections
from shallow to deep layers, which indicates that skip connecting
low-dimensional features might not lead to positive effects. 
Lastly, the performance difference between the best and worst ones
is less than 2.4\% and 1.1\% for the Cornell dataset and the AJD,
respectively, relative to the best performing architecture. Networks
with simple feature extractors, i.e., AlexNet, still achieve
satisifactory results.  Limiting to earlier networks, Resnet-50 backbone
and those in the rows above, the processing speed mostly increases but
with a small decrease in performance; less than 2\% for the Cornell data
set and less than 1\% for the AJD.
The close performance across all models shows that the proposed
grasp keypoint representation is efficient for grasp detection
and robust to network architecture.

{\textbf{Orientation Filter.}}
As shown in the Tables \tabref{bbnets} and \tabref{bbajd}, 
the orientation filter does improve
performance by a small amount over all models.

{\textbf{Heatmap Resolution.}}
Implementing GKNet without the offset mapping and with keypoint output
at the same resolution as the input leads to grasp recognition accuracy
of 94.0\% and 97.77\% for the Cornell dataset and the AJD, respectively.
These values reflect reduced performance relative to the baseline
implementation.  Upsampling the output too much relative to the initial
output is not advisable.
A benefit of the lower resolution heatmaps is a faster inference speed
due to the reduced computation, plus a smaller memory footprint used
during training, even with the offset maps included.

For the remainder of the paper, GKNet will refer to the variant with the
DLA-34 backbone, $R = 4$, and the orientation filter.

\begin {table}[t]
  \centering
  \small
  \caption {GKNet Backbone: Abridged Jacquard Dataset\tablabel{bbajd} }
  \addtolength{\tabcolsep}{-2pt}
  \begin{tabular}{ | l | c | c | c |}
    \hline
    \textbf{Approach} & \textbf{Acc. w o.f.} (\%) & \textbf{Acc. w/o o.f.} (\%) & \textbf{Speed} (fps) 
    \\ \hline
    Alexnet               &     97.37      &     94.53      & 34.48          \\ \hline
    VGG-16                &     98.36      &     96.13      & 21.28          \\ \hline
    Resnet-18             &     97.95      &     95.97      & 31.25          \\ \hline
    Resnet-50             &     98.24      &     95.91      & 25.00          \\ \hline
    Hg-52                 &     97.21      &     93.81      & 15.87          \\ \hline
    Hg-104                &     97.93      &     96.04      & 9.90           \\ \hline
    \textbf{DLA}          & \textbf{98.39} & \textbf{96.99} & \textbf{23.26} \\ \hline
  \end{tabular}
\end {table}

\subsection{Evaluation on the Cornell Dataset \label{secGraspOutCD}}
We first evaluate our work on the Cornell dataset with image-wise and
object-wise splits, following the approach of splitting training and
testing sets described in \cite{chu2018real}. 
The Cornell dataset is older than AJD and has more published performance
outcomes, permitting comparison with a larger set of grasp detection
methods.
In the comparison, we divide state-of-the-art works into 
one-stage approaches, two-stage approaches, 
methods based on grasp quality scores, and other works. 
Table \tabref{eotcd} collects the prediction accuracy for image-wise and
object-wise splittings, as well as inference speed. 
All results are taken from corresponding publications except
those of \cite{morrison2019learning}, which is re-implemented
based on the public repository.
For \cite{chu2018real}, the ResNet-50 results are recorded in Table \tabref{eotcd}.
Most of the published works have a prediction accuracy lower than GKNet,
however their inference time is nearly an order of magnitude larger.
Focusing on the image-wise split, GKNet accuracy is 0.8\% lower than the
best one-stage approach but runs 5 times faster.  While also lower
by 0.8\% than the best two-stage method, it is 10 times faster.  
Similar accuracy statistics hold for the object-wise split.

Some of the backbone networks for these methods are older ones, like
VGG-16 or ResNet-50. The results in Table \tabref{bbnets} indicate that
GKNet's performance continues to best those whose accuracy is below
96.0\%, irrespective of the backbone chosen (so long as it is not
AlexNet). The backbone of \cite{morrison2019learning} is similar to
AlexNet, but does not perform as well as GKNet w/AlexNet backbone.
Thus, the ablation study leading to Table \tabref{bbnets} suggests that
GKNet's good relative performance is more a function of the
representation than the backbone network.

\begin {table}[t] \centering \small \caption {Evaluation on the Cornell
Dataset\tablabel{eotcd} } \setlength{\tabcolsep}{3pt} \begin{tabular}{ |
l | c | c | c |}
    \hline
    {\bf Approach} & \multicolumn{2}{c|}{\bf Accuracy (\%)} & \bf{speed}       \\ \hline
                   & \parbox{0.4in}{\centering image \\ split} 
                   & \parbox{0.4in}{\centering object \\ split} & fps   \\ \hline

    {\bf other work}&                 &                             &      \\ \hline
    \cite{jiang2011efficient}     & 60.5 & 58.3 & 0.02     \\ \hline
    \cite{asif2017rgb}            & 88.2 & 87.5 & --       \\ \hline
    {\bf grasp quality} &                 &                             &      \\ \hline
    \cite{lenz2015deep}           & 73.9 & 75.6 & 0.07     \\ 
\hline
    \cite{mahler2017dex}          & 93.0 & --   & --       \\ 
\hline
    {\bf one-stage} &                 &                             &      \\ \hline
    \cite{morrison2019learning}    & 78.4 & 84.3  &  233.73  \\ 
\hline 
    \cite{wang2016robot}              & 81.8 & --   &  7.10    \\
\hline
    \cite{redmon2015real}         & 88.0 & 87.1 & 3.31     \\ 
\hline
    \cite{kumra2017robotic}        & 89.2 & 88.9 & 16.03    \\ 
\hline
    \cite{guo2017hybrid}          & 93.2 & 89.1 & --       \\ 
\hline
    \cite{zhou2018fully}          & 97.7 & 96.6 & 8.51     \\ 
\hline
    GKNet                & {96.9} & {95.7} & {41.67} \\ \hline
    {\bf two-stage} &                 &                             &      \\ \hline
    \cite{chu2018real}            & 96.0 & 96.1 & 8.33     \\ 
\hline
    \cite{asif2019densely}        & 97.7 & -- & ~4        \\ 
\hline
  \end{tabular}
\end {table}

\begin {table}[t] \centering \small \caption {Backbone Network Comparison\tablabel{bnc} } \setlength{\tabcolsep}{3pt} 
\begin{tabular}{ | l | c | c |}
    \hline
    {\bf Approach} & Network & Parameters \\ \hline

    {\bf other work}              &            &        \\ \hline
    \cite{jiang2011efficient}     & --         & --     \\ \hline
    \cite{asif2017rgb}            & --         & --     \\ \hline
    {\bf grasp quality}           &            &        \\ \hline
    \cite{lenz2015deep}           & --         & --     \\ \hline
    \cite{mahler2017dex}          & --         & 0.18M  \\ \hline
    {\bf one-stage}               &            &        \\ \hline
    \cite{morrison2019learning}   & --         & 0.02M  \\ \hline 
    \cite{wang2016robot}          & --         & --     \\ \hline
    \cite{redmon2015real}         & --         & 0.67M  \\ \hline
    \cite{kumra2017robotic}       & ResNet-50  & 25.56M \\ \hline
    \cite{guo2017hybrid}          & ZF-Net     & 5.30M  \\ \hline
    \cite{zhou2018fully}          & ResNet-101 & 44.55M \\ \hline
    GKNet                         & DLA-34     & 15.74M \\ \hline
    {\bf two-stage}               &            &        \\ \hline
    \cite{chu2018real}            & ResNet-50  & 25.56M \\ \hline
    \cite{asif2019densely}        & DenseNet*  & --     \\ \hline
    
  \end{tabular}
\end {table}

\begin{figure}[t]
    \centering
    \includegraphics[width=\columnwidth,clip=true,trim=10pt 10pt 10pt 10pt]{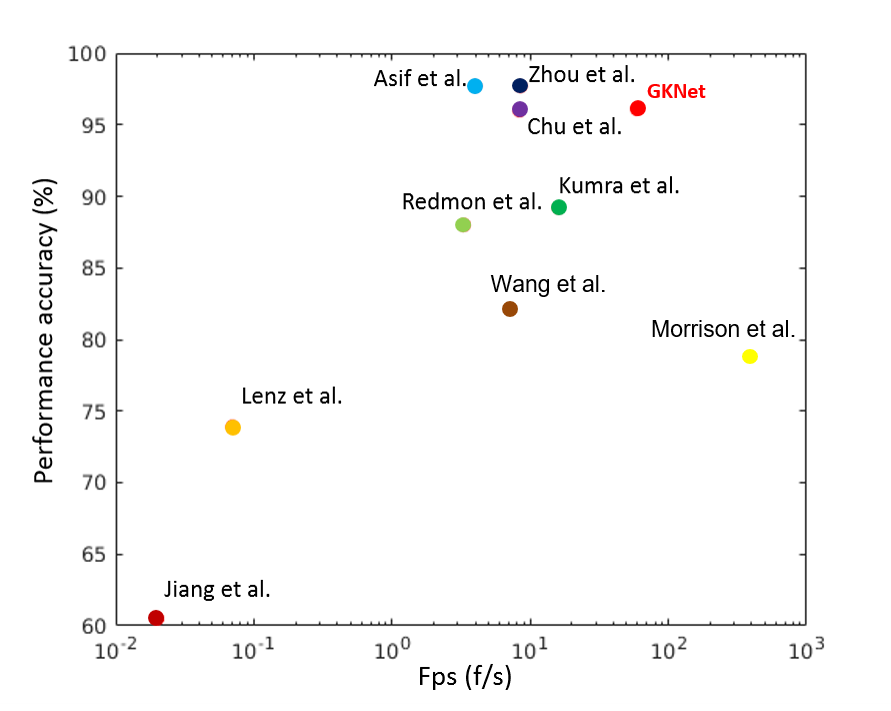}
    \caption{Plot of performance accuracy (\%) versus inference speed
    (fps) using the Cornell benchmark for published works. 
    Performance accuracy is tested on image-wise split.
    Inference speed is in log-space.  Closer to top-right is better.%
    \label{fig_pa_fps}}
\end{figure}

To gain a better sense for how GKNet optimizes for both accuracy and
speed, Fig.~\ref{fig_pa_fps} plots image-wise prediction accuracy versus
inference speed for the ten methods from Table \tabref{eotcd} for which
such information is available. The preferred location for an
implementation that jointly optimizes accuracy and speed will be near
the top-right corner of the figure. GKNet is the approach closest to the
top-right corner, which indicates that it achieves a good trade-off between
prediction accuracy and inference speed.
One factor affecting this conclusion is the different hardware
configurations of the methods.  To address this factor, GKNet was 
run on an NVIDIA GeForce GTX 1070 and achieved 37.04 fps. Its
location in the graph would be close to the plotted GKNet location, which
means that GKNet still provides a favorable trade-off between detection
accuracy and inference speed on less capable hardware. 
Being able to maximize both is advantageous.  
Strong prediction accuracy should translate to high success rate when 
integrated into a manipulation pipeline. 
Later experiments will show that the GKNet achieves top ranked success rates 
in eye-to-hand static grasping experiments relative to 
other published experiments, with a small performance gap between visual
and experimental benchmarking.  
Low inference time supports robust grasping when the robot executes 
tasks in dynamic settings (to be shown in Sec.~\ref{secResGrasp}).

To get a sense for the backbone network architecture of GKNet relative to
existing methods, Table \tabref{bnc} collects the number of trainable
parameters for various backbone networks used in published works.
For those methods which do not employ off-the-shelf standard networks
for extracting features or do not provide detailed network architectures,
corresponding results are left empty. 
For \cite{redmon2015real}, \cite{mahler2017dex}, and
\cite{morrison2019learning} which do not employ standard off-the-shelf
networks, the backbone network parameter counts are computed based on
the network architectures described in the publications.
For \cite{asif2019densely}, their base network is a variant of DenseNet but
the number of parameters isn't provided.  
As shown in Table \tabref{bnc}, DLA-34, which contains fewer parameters
via efficient skip connections structures, contributes to the
state-of-the-art trade-off between accuracy and speed. 
Interestingly, \cite{mahler2017dex} achieves close result with a
backbone network that is 100 times smaller, which indicates that
standard state-of-the-art networks might be too deep and contain
redundant parameters. 

\begin {table}[t]
  \centering
  \vspace*{1.5ex}
  \caption {Evaluation on the Abridged Jacquard Dataset\tablabel{eotjd}}
  \small
  \setlength{\tabcolsep}{3pt}
  \begin{tabular}{ | l | c | c |}
    \hline
    {\bf{Approach}} & \bf{Accuracy (\%)}     & {\bf{Speed (fps)}} \\
      \hline
      {\bf one-stage} & & \\ \hline
      \cite{morrison2019learning}
        & 86.41      &  250.78     \\ \hline 
      GKNet 
        & \bf{98.39} & \bf{23.26} \\ \hline
      {\bf two-stage} & & \\ \hline
      \cite{chu2018real}
        & 95.08      &    7.52    \\ \hline
  \end{tabular}
\end {table}

\subsection{Evaluation on the AJD \label{secGraspOutAJD}}
Benchmarking with the AJD dataset will be more limited since the newer
Jacquard dataset has less published evaluations. 
Comparison methods obtained from publicly released code consist of
a one-stage detector \citep{morrison2019learning} and a two-stage detector 
\citep{chu2018real}. For both compared approaches, we employed the same 
training strategies they released for the Jacquard dataset to train 
on the AJD. 
Processing speed estimates were obtained from executing the
approaches on the same system hardware.
Table \tabref{eotjd} contains the evaluation of outcomes for the two
baseline methods.  GKNet performance remains high and outperforms the
published methods.
The slightly higher performance on the Abridged Jacquard dataset is most
likely due to the dense grasp annotations.
The inference speed is lower but still close to frame-rate. 

\section{Physical Experiments and Evaluation \label{sec:PE}}
This section introduces the experimental setup and evaluation metric
for all physical experiments.
Manipulation experiments will include several different test and evaluation
methods found in the literature or intended to demonstrate functional
characteristics. They are:
(1) static grasping, 
(2) dynamic grasping, and 
(3) grasping at varied camera angles and 
(4) bin picking. 
Static grasping is a simple test requiring precise grasp detection for
isolated objects, and is commonly performed to establish upper bounds on
performance due to the lack of distractors and impediments to grasping.
It also tests robustness to object type when many different objects are
tested as part of the protocol.
Dynamic grasping experiments test the combination of accuracy and speed,
but require some additional outer-loop elements for converting the grasp
detection system into a grasp tracking one. The experiments primarily
test for speed, with the resulting accuracy gap relative to static
grasping also being an indication of grasp detection consistency.
Grasping under varied camera angles tests the robustness with respect to
viewpoint and demonstrates some form of generalization of the algorithm
relative to this factor.
The bin picking experiment is the most challenging of the four since
grasp detection pipelines require additional outer-loop and
post-processing elements to identify a preferred grasp. Unlike grasp
quality networks, which can be trained under these circumstances,
one-stage and two-stage detectors are usually not trained to work in
these settings.
Some grasp quality networks benefit from simulation for amassing very
large ground-truth data, which can be several orders of magnitude larger
than for manually annotated datasets 
{(i.e., DexNet grasp data is three
orders of magnitude more than the Cornell benchmark dataset).}
Conducting this experiment will involve introducing a simple grasp
prioritization scheme through a model-based grasp scoring system.

\subsection{Robotic Platform Setup}
Our robotic platform setup consists of a Kinect Xbox 360 for image acquisition
and a 7 degree-of-freedom custom-made robotic arm for performing
grasping, lifting, and placing tasks.
The camera setup is an eye-to-hand viewpoint. The distance from the
camera center to table is around 1.2m for all experiments.
The custom-made 7 degree-of-freedom robotic arm, shown in Figure
\ref{fig_bp}, employs DYNAMIXEL servo motors as actuators and is
attached with the custom-made parallel-jaw gripper, which can be viewed
in Figure \ref{fig_s_gm} (a).
The gripper width is 17mm and the maximal open width is 200mm.
The maximal depth is 80mm when the gripper closes.

\subsection{Objects Sets for Manipulation Experiments}

\begin{figure*}[t]
  \centering
  \scalebox{0.98}{
  \begin{tikzpicture} [outer sep=0pt, inner sep=0pt]
  \scope[nodes={inner sep=0,outer sep=0}] 
  \node[anchor=north west] (a) at (0in,0in) 
    {\includegraphics[width=4.35cm,clip=true,trim=0in 0.0in 0in 0in]{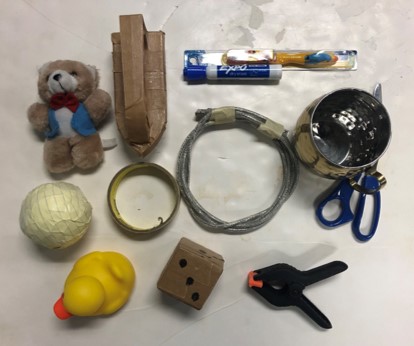}}; 
  \node[anchor=north west, yshift=-0.1cm] (b) at (a.south west)
    {{\includegraphics[width=4.35cm,clip=true,trim=0in 0.0in 0in 0in]{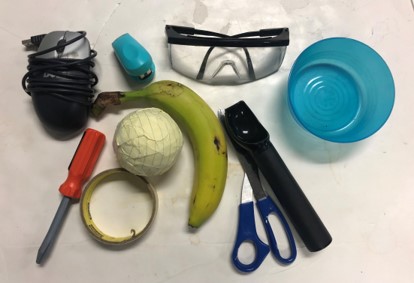}}}; 
  \node[anchor=north west, xshift=0.1cm] (c) at (a.north east)
    {\includegraphics[width=3.95cm,clip=true,trim=0.0in 0.0in 0.0in 0.0in]{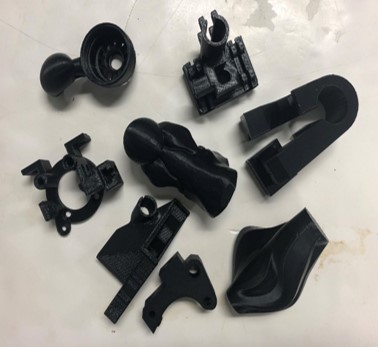}}; 
    \node[anchor=north west, xshift=0.1cm, yshift=-0.1cm] (d) at (a.south east)
    {\includegraphics[width=3.95cm,clip=true,trim=0.0in 0.0in 0.0in 0.0in]{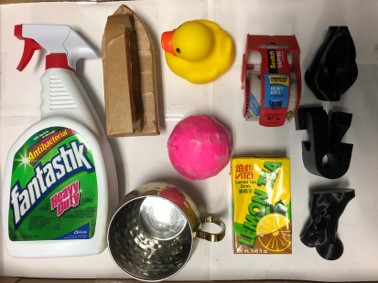}}; 
    \node[anchor=north west, xshift=0.1cm] (e) at (c.north east)
    {\includegraphics[width=8.9cm,clip=true,trim=0.0in 0.0in 0.0in 0.0in]{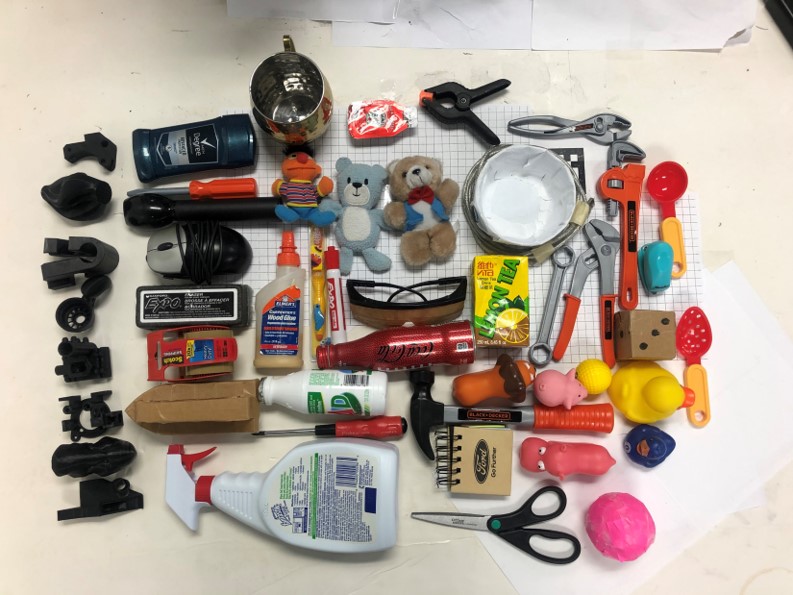}}; 
  \endscope

  \foreach \n in {a,b,c,d,e} 
   {
    \node[anchor=north west,fill=white,draw=white,inner sep=1pt] at (\n.north west) {(\n)};
   }
  \end{tikzpicture}}
  \vspace*{-0.25ex}
  \caption{
    Object sets used for two grasp experiments. (a) The household set (Morrison). (b) The household set (Chu). (c) The adversarial set. (d) The compound set. (e) The bin picking object set.
    \label{fig_obj_set}}
\end{figure*}

Though there are published robotic benchmark object datasets like
the ACRV Picking Benchmark (APB)
and the YCB Object Set  \citep{calli2015benchmarking,leitner2017acrv},
most published works employ their own objects or a subset of objects
from these datasets.  For comparison purposes, we employ 5 object sets
for 4 different kinds of robotic grasping experiments. 
The test sets include novel objects not included in the training datasets. 
They are described below and visualized in Fig. \refsp{fig_obj_set}.
They have some overlap with the APB and YCB sets.

\vspace*{0.25em}
\noindent
{\bf Household Sets.} 
The two different household sets are shown in Fig. \sfigref{fig_obj_set}a 
and \sfigref{fig_obj_set}b, 
as described in \cite{morrison2019learning} and \cite{chu2018real}.
The Morrison set contains 12 common household objects of different sizes
and shapes. The objects were chosen from the standard robotic grasping
datasets \citep{calli2015benchmarking,leitner2017acrv}.
The Chu set contains ten commonly seen objects collected from the
Cornell dataset.

\noindent
{\bf Adversarial Set.} 
The adversarial set, shown in Fig. \sfigref{fig_obj_set}c, consists of eight
3D-printed objects described in Dex-Net 2.0 \citep{mahler2017dex}. 
The objects are characterised by complex and adversarial geometric
features such as smooth and curved surfaces, which increase the
difficulty of grasping. They are meant to undermine grasp scoring
systems. It is less clear what impact these shapes have on other
learning-based grasping approaches.

\noindent
{\bf Compound Set.} The compound set, shown in Fig. \sfigref{fig_obj_set}d,
is composed of seven common household objects and 3 objects from the
adversarial set.

\noindent
{\bf Bin Picking Object Set.} 
The bin picking object set, shown in Fig. \sfigref{fig_obj_set}e, is
collected for bin picking experiments. The object set has 50 objects
consisting of household objects and adversarial objects.

\subsection{Evaluation Metrics for Manipulation Experiments}

Unless otherwise stated, the evaluation criteria for the grasping
experiments will consist of the grasping success rate, which is the
percentage of successful grasps relative to the total amount of grasp
attempts.  A grasp is considered to be a success if the manipulator
grasps, lifts, and holds the object in the air for three seconds.
While some researchers also measure--and advocate for--\textit{mean picks per hour} \citep{mahler2016dex}, we elect to not include this
quantitative measure as it is influenced more by the manipulator
properties, the manipulation code, and the programmed grasping
procedure, than by the visual processing algorithm. In short, within the
perception, planning, and action loop, the planning and action steps
determine the pick rate more so than the perception step. 
This manuscript's scope is related to the perception process, not to the
planning and action processes.

\subsection{Trained Models for Manipulation Experiments}
For the majority of the manipulation experiments, 
the Cornell dataset was the training dataset.  This permits evaluation of 
approaches that trained and tested against Cornell, and also performed
manipulation experiments. Some methods have a performance gap when moving 
from visual processing to embodied grasping, which network training data 
consistency permits comparing with.

The bin picking experiment is the exception, for which GKNet was
trained with the AJD.  Bin picking scenarios involve multiple objects,
for which neither the AJD nor the Cornell dataset has the equivalent
type of training data; both consist of scenes with a single object only.  
The choice to use AJD is due to differences in the annotation sparsity,
or in the ground-truth grasp distributions, between the Cornell dataset
and the AJD.  The denser annotations in the AJD dataset better aid in
locating grasps for object sub-parts, whereas models trained on the
Cornell dataset  tend to provide grasp candidates with respect to the
whole object.  For cluttered environments with overlaid or occluded
objects, the ability to focus on sub-parts improves the ability to
differentiate occlusion geometries and target object regions.

\section{Robot Benchmark Results \label{secResGrasp}}

To confirm that the vision-based outcomes translate to equivalent
manipulation performance for embodied robotic arms, this section
describes the outcomes of the four eye-to-hand manipulation experiments
described in Section \ref{sec:PE}.  Specific details
regarding the experimental setup and testing approach are included, with
comparison to published works where possible.

The aim is to evaluate GKNet with regards to accuracy, speed,
and robustness. Here robustness will be with regards to variation in
object categories, or with respect to a nuisance factor (object
motion, camera viewpoint, clutter). Accuracy is implicitly tested as the
perception component is a limiting factor in grasp success. The current
(top-1) prediction accuracy is sufficiently high (roughly 96\%) that the
experiments primarily test robustness, with processing speed being
evaluated for the dynamic object experiments. A brief discussion
of findings concludes each experimental section. Supplemental Material
is found at \cite{ivagit_GKNet}.

\subsection{Static Grasping}
The static grasping experiment consists of grasping trials for
individual objects. It is a baseline test for understanding expected
performance in ideal situations. 
The test primarily evaluates accuracy of grasping candidate prediction,
and secondly robustness to object variation.
It mirrors the experiments in
\citep{morrison2019learning,  chu2018real}.
\subsubsection{Experimental Setup and Methodology.}
For comparison to published methods whose implementations are publicly
available, the static grasping experiments consist of tests on three
object sets, the Morrison set, the Chu set, and the adversarial set,
each reported individually.  
The choice of the object sets permits comparison with the published
single-object pipeline methods. 
In particular experiments on the Morrison set and adversarial set
compare with \cite{morrison2019learning}, and the Chu set compare with
\cite{chu2018real}.  

An experiment consists of 10 grasp trials for each
object in the chosen object set. 
For a given object set, a single trial begins with an object placed
within the camera's view. 
The manipulation process involves planning a pick sequence, whereby the
manipulator first moves to a location 15cm higher than the chosen grasp
(along the $z$-axis) and oriented downwards, then rotates its wrist to
achieve the predicted grasp angle.  Once in this pre-grasp pose, the
manipulator lowers the gripper to the grasp location and closes. The
gripper is lifted 15cm back up to the pre-grasp pose.
The GKNet grasp candidate selected is the one with the highest
confidence score. 

\begin {table}[t]
  \centering
  \caption {Static Grasping experiments results with 95\% confidence intervals for household set (Morrison)\tablabel{sghsm}}
  \small
  \setlength{\tabcolsep}{4.5pt}
  \begin{tabular}{ | l | c | l | c |}
    \hline
    {\bf object} & \bf{Accuracy} & {\bf object} & \bf{Accuracy}            \\ \hline
    cup          &      9/10     &  ball        & 10/10                    \\ \hline 
    brush        &      9/10     &  toothbrush  &  9/10                    \\ \hline 
    bear toy     &     10/10     &  dice        & 10/10                    \\ \hline 
    tape         &      9/10     &  duck toy    & 10/10                    \\ \hline 
    marker       &     10/10     &  clamper     &  8/10                    \\ \hline 
    screwdriver  &     10/10     &  cable       & 10/10                    \\ \hline\hline
    \multicolumn{2}{|l|}{GKNet}  &   \multicolumn{2}{c|}{95.00 $\pm$ 3.90}     \\ \hline 
    \multicolumn{2}{|l|}{\cite{morrison2019learning}} & \multicolumn{2}{c|}{92.00 $\pm$ 4.85}                   \\ 
\hline
  \end{tabular}
  \vspace*{1.5ex}
  \caption {Static Grasping experiments results with 95\% confidence intervals for household set (Chu)\tablabel{sghsc}}
  \setlength{\tabcolsep}{4.5pt}
  \begin{tabular}{ | l | c | l | c |}
    \hline
    {\bf object} & \bf{Accuracy} & {\bf object} & \bf{Accuracy}            \\ \hline
    banana       &     10/10     &  stapler     & 9/10                     \\ \hline 
    glasses      &     10/10     &  spoon       & 10/10                    \\ \hline 
    ball         &     10/10     &  bowl        & 10/10                    \\ \hline 
    tape         &      9/10     &  scissors    & 9/10                     \\ \hline 
    screwdriver  &     10/10     &  mouse       & 9/10                     \\ \hline\hline
    \multicolumn{2}{|l|}{GKNet}  &   \multicolumn{2}{c|}{96.00 $\pm$ 3.84}   \\ \hline 
    \multicolumn{2}{|l|}{\cite{chu2018real}} & \multicolumn{2}{c|}{86.00 $\pm$ 6.80}     \\ 
\hline
  \end{tabular}
  \vspace*{1.5ex}
  \caption {Static Grasping experiments results with 95\% confidence intervals for adversarial set\tablabel{sgas}}
  \setlength{\tabcolsep}{4.5pt}
  \begin{tabular}{ | l | c | l | c |}
    \hline
    {\bf object} & \bf{Accuracy} & {\bf object}      & \bf{Accuracy}            \\ \hline
    bar\_clamp   &     10/10     &  gearbox          & 10/10                    \\ \hline 
    part1        &      9/10     &  vase             &  9/10                    \\ \hline 
    pawn         &     10/10     &  turbine\_housing & 10/10                    \\ \hline 
    part3        &      8/10     &  nozzle           & 10/10                    \\ \hline\hline
    \multicolumn{2}{|l|}{GKNet}  &     \multicolumn{2}{c|}{95.00 $\pm$ 4.78}  \\ \hline 
    \multicolumn{2}{|l|}{\cite{morrison2019learning}} & \multicolumn{2}{c|}{84.00 $\pm$ 8.03} \\ 
\hline
    \multicolumn{2}{|l|}{\cite{mahler2017dex}} & \multicolumn{2}{c|}{93.00 $\pm$ 5.59} \\ 
\hline
  \end{tabular}
\end {table}

\subsubsection{Analysis of Outcomes.}
Tables \tabref{sghsm}\,, \tabref{sghsc}\,, and \tabref{sgas} collect the
recorded outcomes of the grasping tests.  They also present the published
results of the analogous trials for the corresponding baseline method
\citep{morrison2019learning, chu2018real, mahler2017dex}. The 95\%
confidence intervals (as binary variable) are also computed for
each method in all three tables.
In \cite{chu2018real}, two criteria for grasp selection were explored,
the top-1 confidence and the nearest-to-center approach.  To match the
GKNet selection criteria, the outcome for top-1 confidence is reported
in Table \tabref{sghsc} (86\% vs 89\% for nearest-to-center).
GKNet has the highest success rate across the three experimental
datasets, which indicates that the keypoint representation for grasp
prediction may be an effective means to hypothesize candidate grasps for
unknown objects.  Looking at the adversarial set outcomes, 
GKNet achieves similar but slightly better performance than \cite{mahler2017dex} 
whose approach was trained on the test objects, which indicates the
capability of GKNet to generalize to novel objects with rich interior
geometry.
Additionaly, GKNet outperforms \cite{morrison2019learning} by 11\% and
nearly matches its own performance on the household objects. This parity
in outcomes for household versus adversarial sets suggests that the
keypoint representation is less affected by adversarial examples.
The confidence intervals of GKNet across the three tests indicate that
these performance outcomes have statistical support. 
GKNet outperforms \cite{chu2018real} for the Chu household set 
and \cite{morrison2019learning} for the adversarial set. GKNet achieves
similar performance as \cite{mahler2017dex} for the adversarial set.
GKNet and \cite{morrison2019learning} performance is similar for the
Morrison household set.

When compared to the vision grasp detection accuracy results
from Tables \tabref{eotcd} and \tabref{eotjd}, the performance gap
between the perception only and the perceive-plan-action pipelines is
smallest for GKNet. There is a less than 1\% difference in success rate
from the vision-only experiment to the manipulation experiments. 
In contrast \cite{chu2018real} drops by 10\%, while
\cite{morrison2019learning} performs poorly for vision-only but executes
well. The vision versus manipulation gap would suggest higher variation
in grasp performance outcomes for \cite{morrison2019learning} as more
objects are tested.

\subsubsection{Discussion.}
The results show that GKNet can generate accurate grasp predictions on
isolated static objects that translate well to actual execution for an
embodied robotic arm.  It is capable of predicting reliable grasp
candidates for objects with complex and adversarial shapes. The tests
confirm that GKNet is robust to object shape. The consistency between
vision-only and embodied manipulation grasp accuracy suggests that GKNet
might have improved robustness properties over other RGB-D grasp
identification pipelines. 
Additional evidence for this possibility lies in the next subsection, which
compares grasping outcomes with other published works.

\subsection{Comparison of Grasping Outcomes with Published Work}
To place the performance of GKNet within the greater context of existing
grasping approaches, Table \tabref{gcpw} collects the success rates of
published grasping research efforts. 
All results cited involved static grasping of single
objects located on a tabletop workspace as measured by a single camera
view, with the exception of \cite{viereck2017learning} which employed
eye-in-hand visual servoing.
Since other experimental conditions in the published outcomes varied,
we include in the table two details of the physical grasping
experiments: the number of unique objects tested and the total number of
trials across the objects. 
In a few cases, one or the other quantity was missing. 
The placement of GKNet in the table is intentional, as the
algorithms above it process image data and output 2D grasp
representations,
while the algorithms below it use point cloud information and 3D grasp
representations.
The method in \cite{lin2020using} is a mixed method whose CNN segments
the depth data to support grasp synthesis post-processing on point clouds.

\begin{table}[t]
  \centering
  \caption{Grasping Comparison to Published Works \tablabel{gcpw}}
  \small
  \begin{threeparttable}
    \setlength\tabcolsep{4pt}
    \begin{tabular}{|l|c|c|c|c|}
    \hline
      \multirow{2}{*}{\bf{Approach}} & 
      \multirow{2}{*}{\bf{Year}} & 
      \multicolumn{2}{ c|}{\multirow{2}{*}{\bf{Settings}}} & 
      \bf{Success} \\ 
       & 
       & 
       \multicolumn{2}{ c|}{} 
       &
      \bf{Rate (\%)} \ \   \\ 
    \hline
    & & \multicolumn{1}{l|}{Objects} & 
      \multicolumn{1}{l|}{Trials} & 
      \\ 
    \hline
    \cite{jiang2011efficient} & 2011 
      & 12  & - & 87.9
      \\ \hline
    \cite{lenz2015deep} & 2015 
      & 30  & 100 & 84.0/89.0\tnote{*}\rule{0pt}{2.2ex}
      \\ \hline
    \cite{pinto2016supersizing} & 2016 
      & 15  & 150 & 66.0
      \\ \hline
    \cite{johns2016deep} & 2016 
      & 20  & 100 & 80.3
      \\ \hline

    \cite{watson2017real} & 2017 
      & 10  & -   & 62.0
      \\ \hline
    \cite{mahler2017dex}  & 2017 
      & 10  & 50  & 80.0
      \\ \hline
    \cite{asif2017rgb} & 2017 
      & 35  & 134   & 91.9
      \\ \hline
    \cite{viereck2017learning} & 2017 
      & 10  & 40   & 97.5
      \\ \hline
    \cite{morrison2019learning} & 2018 
      & 20  & 200   & 87.0
      \\ \hline
    \cite{chu2018real} & 2018 
      & 10  & 100   & 89.0
      \\ \hline

    \cite{satish2019policy} & 2019 
      & 8 & 80    & 87.5
      \\ \hline 
      
    \cite{asif2019densely} & 2019 
      & -  & $>$200   &
89.0/90.0\tnote{**}\rule{0pt}{2.2ex}\rule{0pt}{2.2ex}  
      \\ \hline 
    \cite{lu2020planning} & 2020 
      & 10  &  30  & 84.0
      \\ \hline \hline   
    
    GKNet   & - & 30 & 300  & 95.3  $\pm$ 2.4

    \\ \hline \hline
    \cite{kopicki2016one} & 2016
      & 45  & 45  & 77.8
     \\ \hline
    \cite{ten2018using} & 2018
      & 30  & 214  & 85.0  
    \\ \hline
    
    \cite{liang2019pointnetgpd} & 2019 
      & 10  & 100   & 82.0
      \\ \hline 
    \cite{mousavian20196} & 2019 
      & 17 & 51    & 88.0
      \\ \hline 
    \cite{lin2020using} & 2020 
      & 10  &  100  & 94.0

      \\ \hline
    \cite{lou2020learning} & 2020 
      & 5  & 50   & 82.5
      \\ \hline 
    \cite{wu2020grasp} & 2020 
      & 20  &  60  & 85.0

      \\ \hline 
  \end{tabular}
  \begin{tablenotes}
    \footnotesize
    \item[*] Success rate of 84\%\,/\,89\% achieved on Baxter\,/\,PR2 robot.
     \item[**] Success rate of 89\%\,/\,90\% achieved on ResNet\,/\,DenseNet.
  \end{tablenotes}
  \end{threeparttable}
\end{table}

\subsubsection{Discussion of Source Data.}
Across the published works, the grasping approaches were tested with
different object sets, numbers of total trials, and successful grasp
criteria.  
Some of the differences are noted here.
The majority of the listed works perform experiments on common household
objects collected by the respective authors, thus there will be some
differences in the object sets. However, more recent works tend to
obtain their sets from 
existing benchmark datasets \citep{calli2015ycb, leitner2017acrv}, 
or attempt to match those of others like \citep{mahler2017dex}.
In some cases the experiments were not repeated trials from a fixed set
of objects so they will not divide (trials by objects) evenly. 
For evaluating success, almost all works utilize a similar policy: 
the grasp is successful when the object is grasped and lifted in the
air. DexNet 2.0  \cite{mahler2017dex} has a slightly different
evaluation metric which requires the gripper to keep hold of the object
after lifting, transporting, and shaking.
One difference regarding the GKNet and \cite{morrison2019learning} rows is that they aggregate the statistics from 
Tables \tabref{sghsm}\,-\,\tabref{sgas}, meaning that the success rate 
percentages include common household objects and adversarial objects.  
In spite of these differences, rough comparisons can be made since the
experiments involved single objects on a clutter-free tabletop.
Two publications reported grasping outcomes using
multiple views, which are are not included in the table.
\cite{kopicki2016one} achieved 84.4\% with 7 views and \cite{ten2018using} 
improved the performance to 87.8\% with a second depth camera for
recovering a richer, more accurate point cloud. 

\subsubsection{Analysis of Outcomes}
Using the confidence interval of GKNet as a starting point for the
comparison, only three methods score higher than the lower limit of the
confidence interval (92.9\%), 
\cite{viereck2017learning} and \cite{lin2020using}.
The remaining outcomes can be considered to be lower performing grasping
methods presuming that the grasping tests are roughly similar.
For certain object subsets, the methods may perform comparably, but for more
general tasks they will most likely exhibit lower performance.
Returning to \cite{viereck2017learning}, whose success rate is at the
high end of the GKNet confidence interval. Two properties differentiate
the results, with the first being the lower numbers of objects and
trials per object, and the second being the eye-in-hand visual servoing 
algorithm design. The difference in trials lends some statistical
uncertainty to the comparison, while the eye-in-hand nature makes the
algorithm more compatible with the later test of dynamic {\em eye-to-hand}
grasping in Sec. \ref{ss:dynGrasp}, where the performance gap is
smaller; GKNet averages 96.5\% for a 1\% performance gap.
The eye-in-hand approach permits multiple estimates from potentially
different views over time and may reduce the likelihood of a grasp error
based on how the sequential estimates are generated and processed.
Additionally, the tests of different viewpoints, Sec. \ref{ss:camAngle},
suggest that there may be a view capable of consistently matching the
performance of \cite{viereck2017learning} from a single image.  The best
performing viewing angle scores 97\% for a 0.5\% performance gap.  There is
also value to exploring what additional, complementary implementation
details from \cite{viereck2017learning} might translate to GKNet: unlike
GKNet, \cite{viereck2017learning} is a grasp sampling and scoring mechanism.
Thus, rather than elect the top-1 grasp candidate, it might be useful to
score and select from amongst the top candidates. The findings in
\cite{satish2019policy} suggests that this would be of value, though the
outcomes of \cite{satish2019policy} for individual object grasping do not
reflect it. 


Lastly for \cite{lin2020using}, whose performance is just below that
of GKNet, the grasping approach prioritizes the detection of shape
primitives to inform grasp hypothesis generation and scoring. 
It consists of a deep learning primitive segmentation step followed by a
more traditional, and slower, model-based grasping pipeline.  The
pipeline generates 3D grasp candidates and scores them to output a
single grasp to execute. Given the strong performance of this system,
there is a value to more explicitly encoding shape into the network
structure or processing pipeline, followed by a GKNet form of grasp
recognition (with grasp re-ranking). 

\subsubsection{Discussion.}
In summary, GKNet represents a top performing grasp pipeline with a
different and complementary network and processing structure to 
comparably performing systems. GKNet's processing time is amongst the
fastest of existing methods.  The tests span a diverse set of objects,
such that the high success rate demonstrates consistency of grasp
performance relative to variation in object shape.  This provides
evidence for the robustness of GKNet to object shape or form, for
graspable objects whose size is comparable to the gripper opening
distance.
The strength of GKNet lies in the keypoint representation, which should
be studied further to understand how it may naturally integrate with
explicit primitive shape modeling and grasp scoring methods, and how it
may extend to $SE(3)$ grasps.

\subsection{Grasping at Varied Camera Angles \label{ss:camAngle}}
The Cornell and AJD datasets are created from annotating imagery with
similar camera views, effectively a top-down view. 
Consequently, training of GKNet uses imagery with a top-down camera
perspective, which need not be the configured perspective for a
manipulation setup.  In such cases, it is important for the grasping
algorithm to provide consistent grasping performance insensitive to
deviations from the top-down perspective. The experiment described in
this section evaluates static grasping performance consistency to varied
camera angles.

\subsubsection{Experimental Setup and Methodology.}
\begin{figure}[t]
  \centering
  {\includegraphics[width=0.9\columnwidth,clip=true,trim=0in 0in 1in 1.2in]{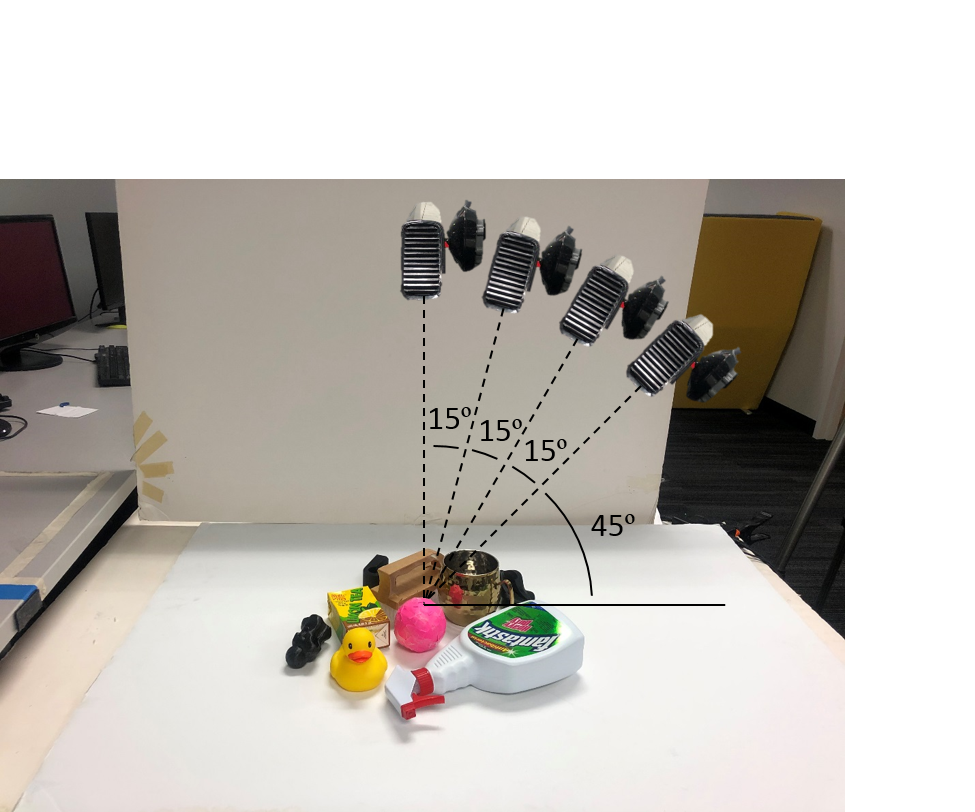}}
  \caption{Experimental setup for grasping at varied camera angles (side
  view). 
  \label{fig_exp3_ill}}
\end{figure}

The setup and methodology is similar to the static grasping experiments,
with camera viewing angle as the independent variable.
The view angles vary from 90\degree to 45\degree in 15\degree
decrements such that the distance from the object to the camera
frame is consistent across the experiments, see Fig. \ref{fig_exp3_ill}\,.
The object set used is the compound object set, Fig.~\ref{fig_obj_set}(d).
A top-down grasp is commanded to pick each object.
There are 10 grasp trials per object with randomized poses for each trial.
The view angle does not go below 45\degree because the data pre-processing
(\S \ref{secDataTrain}\,) approach is incompatible with learning to grasp
on highly skewed background planes, as occurs for table-like support
surfaces viewed at low incidence angles.  

\subsubsection{Analysis of Outcomes.}
As shown in Table \tabref{gvcp}, GKNet achieves grasp success rate of
95.00\%, 97.00\%, 94.00\% and 93.00\% 
for the camera orientation of {90\degree},
{75\degree}, {60\degree} and {45\degree}, respectively. 
Among the results, the experiment of {75\degree} camera angle achieves the best 
outcome, while the outcome of {45\degree} camera angle is the worst. 
Interestingly, the {75\degree} success rate lies quite close to that of
\cite{viereck2017learning} from Table \tabref{gcpw}\,, indicating that 
GKNet could possibly be configured to achieve state-of-the-art performance. 

Choosing the outcome of 75\degree as the reference, outcomes of
{90\degree} and {60\degree} fall inside its 95\% confidence interval
(binary variable) while the outcome of {45\degree} falls just
outside.  
The {45\degree} viewpoint sufficiently differs in image structure that 
network inference begins to degrade, albeit by a small amount as 
indicated by the fact that the {90\degree}, {60\degree}, and {45\degree} 
outcomes are close to each other. 
Looking at Table \tabref{gcpw}\,, all of the views continue to outperform
the majority of the other methods when considering the lower boundaries of
the intervals. 
The approach whose outcome lies within the confidence intervals of
the {90\degree}, {60\degree}, and {45\degree} cases is \cite{asif2017rgb}. 
For the {45\degree} case, three methods lie within the confidence
interval: \cite{asif2019densely,chu2018real,lenz2015deep}.
In all cases, GKNet  mean values are higher and the methods are at the
lower end of the intervals.  
The {45\degree} viewpoint reflects a significant view change given that 
GKNet was designed under the assumption of consistent background depth values, 
which a {45\degree} viewing angle does not have. 
The difficulty is reflected in the lower
success rate and larger confidence interval.  

\subsubsection{Discussion.}
These aggregate outcomes indicate that GKNet is relatively robust to
camera viewing angle (up to 45$^\circ$ as tested). It will work well for
camera configurations that deviate from top-down views, which commonly
happens when configuring a camera to operate as an eye-to-hand
manipulation system.  The lower performance at the more oblique angle
(45$^\circ$) may result from the loss of top-down geometry impacting
top-down grasping. In these cases, approaching from an angle more
consistent with the viewpoint should improve performance.
More extreme camera angles ($30^\circ$ or less) would benefit from
estimation and isolation of the dominant plane to remove its impact, 
from generalizing grasps to not require top-down views, or from
pursuing $SE(3)$ grasp recognition extensions.

\begin {table}[t]
  \centering
  \caption {Results of grasping at varied camera angles experiments with 95\% confidence intervals for the compound set.\tablabel{gvcp}}
  \setlength{\tabcolsep}{4.5pt}
  \small
  \begin{tabular}{ | l | c | c | c | c |}
    \hline
    \multirow{2}{*}{\bf{object}} & \multicolumn{4}{c|}{\bf{Accuracy (\%)}}           \\ \cline{2-5}
             & 90\degree & 75\degree & 60\degree & 45\degree \\ \hline 
 bar\_clamp  &  10/10    &  10/10    &  10/10    &  10/10    \\ \hline 
 pawn        &  10/10    &  10/10    &   9/10    &  10/10    \\ \hline 
 vase        &   8/10    &   9/10    &   9/10    &   9/10    \\ \hline
 tape        &  10/10    &  10/10    &  10/10    &  10/10    \\ \hline
 sprayer     &  10/10    &  10/10    &   9/10    &   9/10    \\ \hline
 juice box   &  10/10    &  10/10    &  10/10    &  10/10    \\ \hline
 duck toy    &   9/10    &  10/10    &   9/10    &   9/10    \\ \hline
 cup         &   9/10    &   9/10    &   9/10    &   8/10    \\ \hline
 brush       &  10/10    &   9/10    &  10/10    &  10/10    \\ \hline
 ball        &   9/10    &  10/10    &   9/10    &   8/10    \\ \hline \hline 
 average (\%) &  95.00  &  97.00 &  94.00  &  93.00  \\
 95\% interval & $\pm$4.27 & $\pm$3.34 & $\pm$2.37 & $\pm$5.00 \\ \hline      
  \end{tabular}
\end {table}

\subsection{Dynamic Grasping \label{ss:dynGrasp}}

An additional nuisance factor that may occur during manipulator
deployment is object movement in a consistent direction, such as when
grasping from a conveyor belt or other consistently moving support
surface. 
A dynamic grasping experiment following the protocol 
of \cite{morrison2019learning} tests real-time robustness to object
movement.

\subsubsection{Implementation.}
To evaluate GKNet in real-time tasks, we perform dynamic grasping
by implementing a Position Based Visual Servoing (PBVS) 
controller \citep{kragic2002survey}.  PBVS control requires real-time
tracking of the robot pose to compute the error between the current and
desired pose.
For the eye-to-hand camera view of the grasping setup, the Depth-Based Robot 
Tracking (DBRT) algorithm \citep{cifuentes2016probabilistic} estimates the
current pose of the robot from a depth camera stream. 
The depth image stream is generated at 30Hz by a Kinect Xbox 360. 
A simple detection-based tracking method informs grasp selection.
At the begining of each round, GKNet outputs the grasp detection with
the highest confidence score and the robot is controlled by the PBVS
controller to move accordingly. 
Since the object is expected to move, future processing from GKNet extracts
the top 5 grasp candidates for each frame and compares these five
predictions with the selected one at the previous frame. 
The grasp candidate with the lowest Euclidean distance to the previous
grasp selection is chosen as the grasp to achieve, if it is below a
tolerance threshold, $\tau_{\rm close}$.  If no grasp passes the
threshold, then the previous grasp pose remains the target grasp pose,
under the presumption that the gripper has occluded the object. 
Closed-loop control will be stopped after the robot reaches the
predicted position and is about to grasp the object.  Upon reaching the
target pose, the gripper closes to grab the object, then lifts.

\subsubsection{Experimental Setup and Methodology.}
The setup and methodology is similar to the static grasping experiments,
however the target object is moved by the experimenter with a
translation of at least 100$mm$ and a rotation of at least {25\degree}
about the object's center during each individual trial.
The direction of travel is linear and randomly chosen, as is the
rotation, and is effected by moving a piece of paper upon which the
object is placed.  The movement is a slow continuous action beginning
right after the manipulator starts to move.  The entire movement lasts
for 3 seconds.  There are 10 grasp trials for each object in the
household set (Morrison) and the adversarial set, for 200 grasp trials
in total.  The grasp attempt is rated successful if the object is
lifted.

\begin {table}[t]
  \centering
  \caption {Results of dynamic grasping experiments with 95\% confidence intervals for household set (Morrison)\tablabel{dghsm}}
  \setlength{\tabcolsep}{4.5pt}
  \small
  \begin{tabular}{ | l | c | l | c |}
    \hline
    {\bf object} & \bf{Accuracy} & {\bf object} & \bf{Accuracy}            \\ \hline
    cup          &     10/10     &  ball        & 9/10                     \\ \hline 
    brush        &      9/10     &  toothbrush  & 10/10                    \\ \hline 
    bear toy     &     10/10     &  dice        & 10/10                    \\ \hline 
    tape         &      9/10     &  duck toy    & 10/10                    \\ \hline 
    marker       &     10/10     &  clamper     &  9/10                    \\ \hline 
    screwdriver  &     10/10     &  cable       & 10/10                    \\ \hline\hline
    \multicolumn{2}{|l|}{GKNet}  &     \multicolumn{2}{c|}{96.67 $\pm$ 3.21} \\ \hline 
    \multicolumn{2}{|l|}{\cite{morrison2019learning}} & \multicolumn{2}{c|}{88.00 $\pm$ 5.81} \\ \hline
  \end{tabular}
  \centering
  \vspace*{1.5ex}
  \caption {Results of dynamic grasping with 95\% confidence intervals for adversarial set\tablabel{dgas}}
  \setlength{\tabcolsep}{4.5pt}
  \small
  \begin{tabular}{ | l | c | l | c |}
    \hline
    {\bf object} & \bf{Accuracy} & {\bf object}      & \bf{Accuracy}            \\ \hline
    bar\_clamp   &     10/10     &  gearbox          & 10/10                    \\ \hline 
    part1        &      8/10     &  vase             &  9/10                    \\ \hline 
    pawn         &     10/10     &  turbine\_housing & 10/10                    \\ \hline 
    part3        &     10/10     &  nozzle           & 10/10                    \\ \hline\hline
    \multicolumn{2}{|l|}{GKNet}  &     \multicolumn{2}{c|}{96.25 $\pm$ 4.16}  \\ \hline 
    \multicolumn{2}{|l|}{\cite{morrison2019learning}} & \multicolumn{2}{c|}{83.00 $\pm$ 8.23}  \\ \hline
  \end{tabular}
\end {table}

\subsubsection{Analysis of Outcomes.}
Tables \tabref{dghsm} and \tabref{dgas} report the individual trial
outcomes for the object and the success rate. 
GKNet achieves 96.67\% and 96.25\% on the household set (Morrison) and
adversarial set. 
These success rates outperform \cite{morrison2019learning} by
8.67\% and 13.25\%, which is greater than GKNet's performance advantage
in the static grasping experiments. Meanwhile the outcomes of
\cite{morrison2019learning} fall below and outside of the GKNet's
confidence intervals (binary variable) providing additional
support that GKNet outperforms the baseline.

\subsubsection{Discussion.}
Qualitatively, the outcomes demonstrate that GKNet processing
speed and grasp candidate accuracy are both sufficiently good to permit
grasping of slowly continuously moving objects, when combined with a
compatible tracking strategy.  
Quantitatively, the similar average and confidence intervals of the
dynamic grasping success rates to those of the static grasping success
rates confirms robustness of grasp candidate generation to movement.  
GKNet's competitive and consistent performance versus the baseline
provides further evidence that a keypoint approach to grasp recognition
is more beneficial relative to other representations for parallel-plate
type grippers, or grippers functionally similar to them.

\begin{figure*}[t]
    \centering
  \begin{tikzpicture} [outer sep=0pt, inner sep=0pt]
    \scope[nodes={inner sep=0,outer sep=0}] 
    \node[anchor=south west] (a) at (0in,0in)
      {\includegraphics[width=0.97\columnwidth]{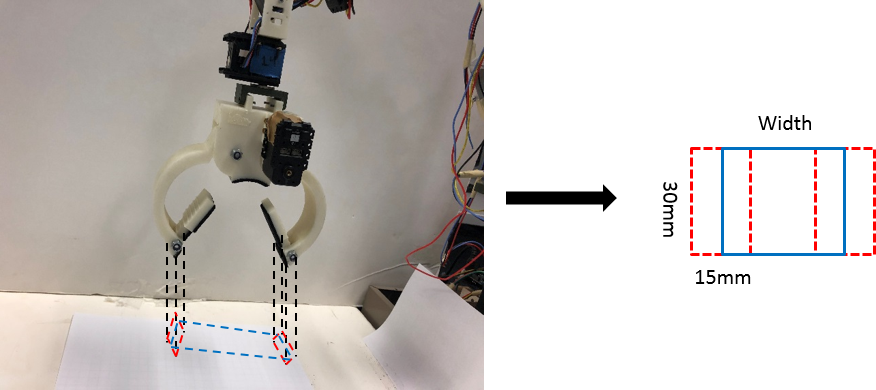}};

  \node[anchor=west, xshift=0.2cm] (b) at (a.east)
    {\includegraphics[width=4.25cm,clip=true,trim=0in 0in 0in 0in]{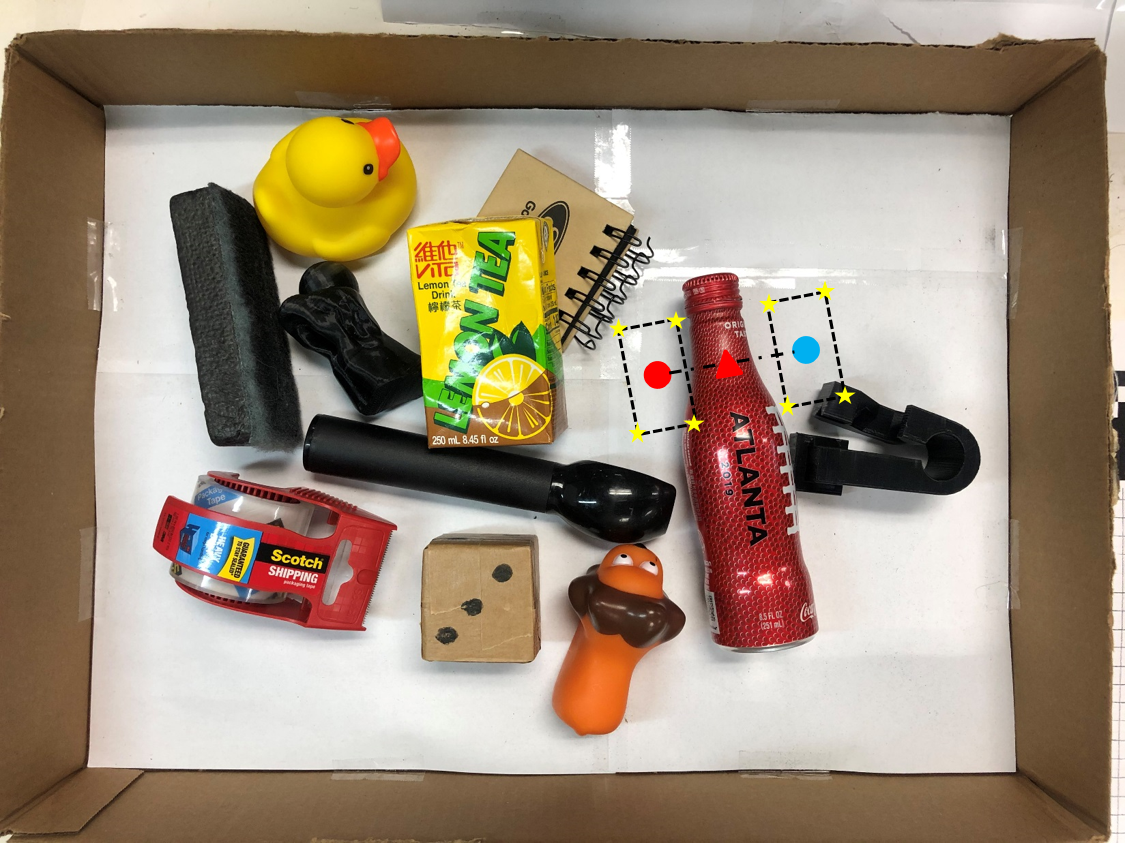}}; 
  \node[anchor=south west, xshift=0.2cm] (c) at (b.south east)
    {\includegraphics[width=4.25cm,clip=true,trim=0in 0in 0in 0in]{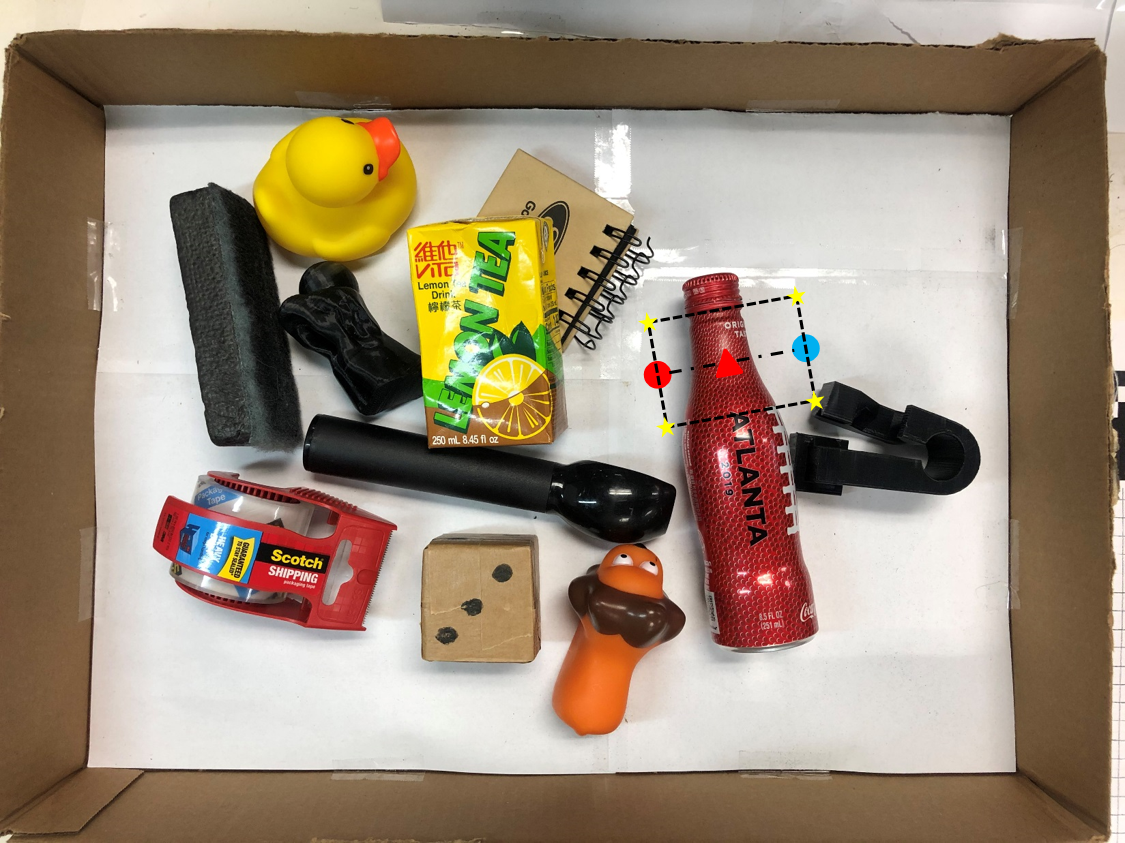}}; 
  \endscope
  \foreach \n in {a,b,c} { 
    \node[anchor=north west,fill=white,inner sep=2pt,outer sep=0pt] at (\n.north west) {(\n)};
  }
  \end{tikzpicture}
  \caption{(a) Simplified 2D gripper model. Red dotted boxes represent the
    two sides of gripper and blue solid box represents the area within
    the gripper. \label{fig_s_gm}
    The gripper model informs the grasp scoring contributions for
    (b) collisions, and (c) occupancy.  \label{fig_gsa}}
\end{figure*}

\subsection{Bin Picking}

The last experiment to perform is a bin picking experiment similar to
\cite{mahler2017learning} and \cite{morrison2019learning}.  
The experiment evaluates robustness of
grasp recognition to environmental clutter, which would thereby examine
the generalization capabilities of GKNet since the network is trained
with images containing only individual objects. 
Multi-object and cluttered environments are scenarios anticipated to be
experienced for some real-world deployments. Thus providing reliable
grasp prediction to other sensed objects is an essential capability of
grasp detection approaches.

\subsubsection{Implementation.}
As a grasp candidate generation strategy, GKNet does not provide a means
to identify which candidate grasp is actually the best to execute in
cluttered settings.  
Grasping objects in cluttered scenes usually involves grasp scoring
algorithms that take into consideration potential collisions and
occlusion geometry
\citep{fischinger2012empty,fischinger2013learning,domae2014fast,ten2017grasp}.
Unlike deep learning grasp quality scoring methods specifically trained
for this problem \citep{mahler2017learning}, GKNet requires additional
post-processing to operate in clutter. 
The additional post-processing is a grasp quality scoring function
designed based on geometric and heuristic approaches
\citep{gualtieri2016high,ten2017grasp}.  
The grasp scoring algorithm implemented for this experiment takes a single
top-down view depth image and a set of grasp detections as input, then
outputs scores for each grasp detection with respect to collision,
occupancy, and grasp height, as described below.

The collision score evaluates the potential for collision by the
two ``fingers'' at the chosen grasp pose.
Given the top-down view depth image, scoring will involve a simplified
model of the 3D gripper into a 2D projected model as shown in
Fig. \ref{fig_s_gm}\,(a).
Two projected red rectangles represent the two sides of gripper, with a
third blue rectangle representing the open area within the two fingers.
For collision checking, per Fig. \ref{fig_gsa}\,(b), the algorithm
processes the pixels inside the rectangles and computes the ratio of
pixels whose depth is larger than the depth of the center keypoint.  
Denote the union of pixel elements in the two rectangular regions by
$\gripRegion$, then the collision score is
\begin{equation} \label{cs}
  s_c = \frac{1}{\left|{\gripRegion}\right|} 
                \sum_{p \in \gripRegion} \Hside(d(p)-d(p_c)),
\end{equation}
where $\Hside(\cdot)$ is the Heaviside function, $p_c$ is the center
keypoint, and $d(\cdot)$ is the depth. The collision score ranges
between 1 (no collisions) to 0 (all collisions).

The occupancy score computes the number of pixels whose depth is
higher the support surface.
For occupancy checking, per Fig. \ref{fig_gsa}\,(c), the main region of
interest lies within the grippers where a hypothesized object part to
grasp lies.  Denote this gripper interior region by $\objRegion$. 
The occupancy score compares the depth values to the nominal depth of
the table or supporting surface at the center keypoint,
\begin{equation} \label{os}
  s_o = \frac{1}{\left|{\objRegion}\right|} 
                                 \sum_{p \in \objRegion} \Hside(d_S(p_c)-d(p)),
\end{equation}
where $d_S(\cdot)$ is the depth map for the supporting surface. The
occupancy score varies from 0 to 1.

The last score involved in the composite grasp scoring function is
the grasp height score, which prioritizes grasping objects from
the top of the heap.
It is computed as the normalized vertical
height off the surface for the hypothesized object located at the 
center keypoint:
\begin{equation} \label{hs}
  s_h = \frac{\left | d(p_c) - d_S(p_c) \right |}{\left | d_S(p_c) \right |}.
\end{equation}
Its range is between 0 and 1, so that the outputs of all three scores
lie within the same range.  The final grasp score is the sum of the scores,
  $s_g = s_c + s_o + s_h$.

\subsubsection{Experimental Setup and Methodology. \subsubseclabel{method}}
\begin{figure}[t]
  \centering
  \vspace*{0.08in}
  \begin{tikzpicture}[inner sep=0pt,outer sep=0pt]
  \node (SF) at (0in,0in)
    {\includegraphics[width=1\columnwidth]{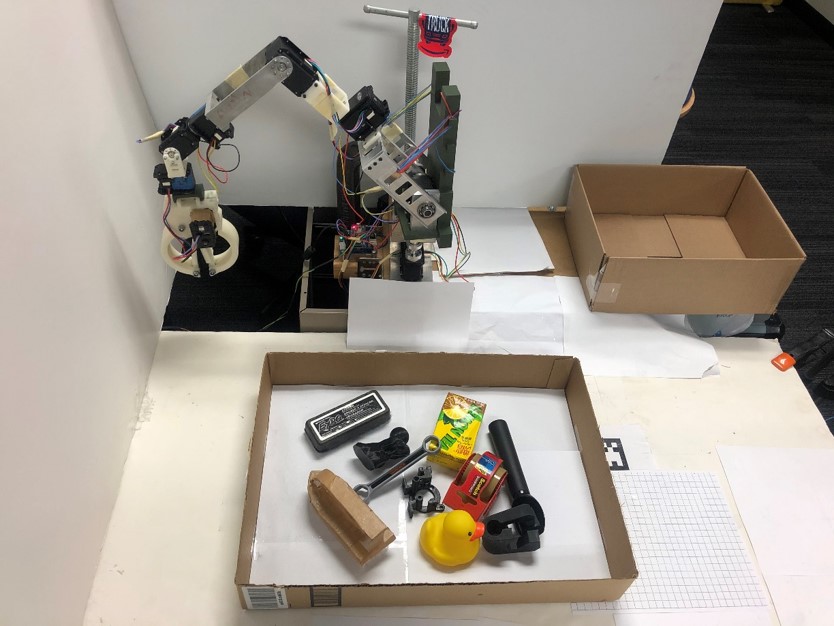}};

  \node[anchor=north west,xshift=30pt,yshift=-160pt,fill=white,fill
  opacity=.5,text opacity=1,inner sep=3pt] at (SF.north west) {\itshape pick bin};
  \node[anchor=north west,xshift=186pt,yshift=-80pt,fill=white,fill 
    opacity=.5,text opacity=1,inner sep=3pt] at (SF.north west) {\itshape place bin};
    
  \end{tikzpicture}
  \caption{Setup of the bin picking experiment. Objects placed into the
    {\em pick bin} should be moved to the {\em place bin}
    \label{fig_bp}}
    \vspace*{0.08in}
\end{figure}

The setup of the bin picking experiment is shown in Fig. \ref{fig_bp}\,. 
The experiment itself follows the protocol in 
\citep{mahler2017learning,morrison2019learning}, and is outlined in 
Algorithm \ref{alg:bin_picking}.
The grasp attempt of an object in the {\em pick bin} is labeled
successful if the robot pick and place action leads to the object being
placed in the {\em place bin}.
The entire experiment iterates until 
\textbf{(a)} there are no objects left in the bin or 
\textbf{(b)} the robot fails to grasp the same object 5 consecutive times.
The bin picking object set, Fig. \ref{fig_obj_set}\,(e), is used.

\paragraph{Trial Configurations and Evaluation.}
The bin picking experiment is performed under two different configurations.  
In one, the number of objects is $N = 5$ and the number of
trials is 20. In the second, the number of objects is $N = 10$
and the number of trials is 10.
In addition to the grasp success rate (SR), a second evaluation metric
is the percent cleared (PC).  The percent cleared is the fraction of
objects that were moved to the {\em place bin}. 
Confident intervals will treat the SR and PC values as continuous
variables.

\paragraph{Baseline Methods and Comparative Evaluation.}
The three comparative baseline methods and scores come from
\cite{mahler2017learning}. They are an image-based Force Closure method,
and the DexNet 2.0 and DexNet 2.1 ($\epsilon = 0.9$) methods.
The Force Closure method performs a random planar force grasp
with friction coefficient $\mu = 0.8$ computed from edge
detection in depth images. It also employs a model-based grasp
scoring system based on \citep{ten2018using, gualtieri2016high}.
Dex-Net 2.0 employs the GQ-CNN model to rank grasps, while
Dex-Net 2.1 ranks grasps using the Dex-Net 2.1 classifier.

The experimental implementation in \cite{mahler2017learning} is not
exactly the same as the one performed here. There are two main
experiments that are closest to the current ones. They are labelled (S1)
and (S2), based on the object set.   The object set (S1) consists of
25 opaque and rigid objects, where the number of trials was set to 20
and 10 for $N=5$ and $N=10$, respectively.  The second object set (S2)
consists of 50 objects, with the additional 25 being transparent,
hinged, and deformable objects.  There were 5 trials per object.

The bin picking object set used here is similar to the set with full
50 objects used in \cite{mahler2017learning} per their description, 
but lying between (S1) and (S2) in difficulty.  
Results from (S1) and (S2) thus establish
nominal upper and lower bounds that would be considered good if the
performance of GKNet with simple grasp quality scoring were to lie near
to or within them.  Force Closure has an established gap versus learning
methods \citep{mahler2017learning}.  GKNet should perform better than
the Force Closure method since it combines a learning-based grasp hypothesis
strategy with model-based grasp scoring. In essence, it replaces the
randomized grasp sampling.

\begin{algorithm}[t]
\caption{Bin Picking Experiment}
\label{alg:bin_picking}
  \begin{algorithmic}[1]
    \State Randomly sample $N$ objects from the object set
    \State Place sampled objects with random poses into the bin
    \While{NOT (condition a OR condition b)}
        \State{GKNet generates Top-100 grasp detections}
        \State{Run grasp scoring algorithm and select the one with highest score}
        \State{The robot attempts to grasp}
        \If{The robot achieves successful grasp}
            \State{The robot places the object into a bin}
        \Else
            \State{Continue}
        \EndIf
    \EndWhile
  \end{algorithmic}
\end{algorithm}

\begin{table}[t]
  \caption{Bin Picking Experiment \tablabel{bpe}}
  \small
  \centering
  \begin{threeparttable}
    \setlength\tabcolsep{4pt}
    \begin{tabular}{|c|c|c|c|c|}
    \hline
      \multirow{2}{*}{\bf{Approach}} & 
      \multicolumn{2}{c|}{\bf{N = 5}} & 
      \multicolumn{2}{c|}{\bf{N = 10}} \\ \cline{2-5}
       & \bf{SR} \tnote{*}\rule{0pt}{2.2ex} \ \
       & \bf{PC} \tnote{**}\rule{0pt}{2.2ex} \ \
       & \bf{SR}
       & \bf{PC}    \\ \hline
  GKNet & 92.1  & 100.0  & 72.1 & 98.0  \\ \hline \hline
  Force Closure (S1) & 54.0 & 97.0 & 55.0 & 92.0 
    \\ \hline
  Dex-Net 2.0 (S1) & 92.0 & 100.0 & 83.0 & 98.0 
    \\ \hline 
  Dex-Net 2.1 (S1) & 94.0 & 100.0 & 89.0 & 100.0 
    \\ \hline \hline
  Force Closure (S2) & - & - & 50.0 & 77.0 
    \\ \hline
  Dex-Net 2.0 (S2) &  -  &  - & 81.0 & 98.0 
    \\ \hline
  Dex-Net 2.1 (S2) &  -  &  - & 85.0 & 100.0 
    \\ \hline
  \end{tabular}
  \begin{tablenotes}
    \footnotesize
    \item[*] SR = Success Rate (\%).
     \item[**] PC = Percent Cleared (\%).
  \end{tablenotes}
  \end{threeparttable}
\end{table}

\color{black}

\subsubsection{Analysis of Outcomes.}
Table \tabref{bpe} collects the outcomes of the bin picking experiment.
For the 5 object setting, the 20 trials led to a 92.1\% grasp success 
rate and 100\% of objects cleared. 
For the 10 object setting, the 10 trials led to a 72.1\% grasp success rate 
and 98\% of objects cleared. 
Reviewing the results relative to the baselines indicates that GKNet plus
simple grasp quality scoring does indeed have the desired properties
described in Section \subsubsecref{method}.
For the $N=5$ case it outperforms Force Closure (S1) and nearly matches 
Dex-Net 2.0 (S1) and Dex-Net 2.1 (S1). 
Recall that (S1) is qualitatively an easier set relative to the 
bin picking set, and GKNet is trained only with individual objects, 
thus one would anticipate higher performance for Dex-Net variants.  
The 95\% confidence interval for GKNet is $92.1 \pm 4.8$, which is above
the Force Closure and contains the DexNet variants.%
The interval supports the conclusions that
GKNet outperforms Force Closure and is comparable to the DexNet variants
for this scenario. 
The comparable performance assessment is further
supported by the 100\% bin-picking \textit{percent cleared} rates for
GKNet and the DexNet variants. For the $N=10$ case there is a larger
difference for the methods, with GKNet performing between Force Closure
and the DexNet variants in terms of success rate. 
The percent cleared matches the Dex-Net 2.0 (S1) and (S2) outcomes.
The 95\% confidence interval of GKNet is $98.0 \pm 2.9$.

For the set (S2) and the $N=10$ case, the GKNet solution continues to
outperform Force Closure with a percent cleared performance close to,
but below, the DexNet variants. 
The success rate lies between Force
Closure and the DexNet variants, but closer to the DexNet variants than
to Force Closure. 
The confidence interval of $72.1 \pm 5.4$ continues to support the
conclusion that GKNet improves upon Force Closure, but in this scenario
does not match the performance of the Dex-Net variants.
As with (S1), however, the percentage of bin-picking trials cleared again 
suggests comparable performance to the DexNet 2.0 variant for the task. 
Since Dex-Net is a grasp scoring CNN trained explicitly to handle 
cluttered scenes, it should have an advantage over model-based grasp 
quality scoring systems. 
The recent effort \citep{satish2019policy} suggests that combining
convolutional network predictions with Dex-Net would give strong
performance, since one limitation of Dex-Net is the sampling step.
Future work should explore the performance of bin picking using a
combined GKNet + Dex-Net method.

The failure cases split into visual and execution errors. 
On the visual side, the employed scoring algorithm prioritizes grasps
whose fingers have less chance to collide with other objects. 
In cluttered environments such grasps may be at the corners or edges of
objects, which do not lead to reliable grasp closure.
Secondly, GKNet sometimes treats multiple objects as one entity and provide
grasp keypoint pairings that spans them, as though trying to grab
multiple objects at once. In the end, the manipulator grabs none of
them. 
Most execution failures were caused by the design of the gripper.
The gripper on the manipulator is not a true parallel plate gripper, but
is a jaw-like gripper.  The volume of space traversed during grasp
closing is a bit larger than the true parallel-plate gripper.  
The simple grasp quality scoring system does not factor this design detail, 
which becomes a more critical factor as the clutter level increases. 
Employing an imitation learning or reinforcement learning grasp 
quality scoring system like Dex-Net for handling clutter should 
address this hard to model aspect.
There were unique errors for the $N = 10$ experiment that prevented full
clearance of the objects.
GKNet plus simple grasp quality scoring has problems selecting accurate
grasp estimates for the small ketchup container, top-middle of Fig.
\ref{fig_obj_set}(e), and for the blue plush bear, just below the
ketchup in Fig. \ref{fig_obj_set}(e).
Of all the grasp candidates, it would select the same failed grasp
candidate, such that a single failure would necessarily imply 5
sequential failures. 
These two objects are the major source of performance degradation. 
Providing a means to recognize failed grasp attempts and choose
alternative grasp candidates would be one way to correct the sequential
mistakes.%

\subsubsection{Discussion.}
The GKNet bin-picking results show that GKNet is good at hypothesizing
grasp candidates, while its use with a model-based grasp scoring and
prioritization system could match learning based methods (i.e., Dex-Net
grasp quality scoring) for low clutter picking scenarios.  There was
some degradation in success rate for more cluttered scenarios, which
impacted the clearance rate.  Further study is needed to understand how
to bridge the light clutter to heavy clutter gap.  Incorporating grasp
quality neural network (GQ-CNN) designs as the grasp scoring mechanism
should be one means to do so, especially if the GQ-CNN models were
trained with diverse clutter scenarios.

\subsubsection{Comparison to 6-DoF Grasping Methods.}
Grasping objects in clutter is a commonly conducted experiment for 6 DoF
grasping methods \citep{murali20206, sundermeyer2021contact,
ten2017grasp,
lou2020learning, liang2019pointnetgpd, ni2020pointnet, zhao2020regnet,
qin2020s4g, ten2018using}. 
The goal of clutter removal is to move the objects somewhere else, which
is similar to bin picking.
Across these published works, experiments were conducted under similar
protocols but can differ in the object set, number of objects placed in
the scene, number of total runs, and stopping criteria of each run. 
Therefore, these results are not put into Table \tabref{bpe} but
discussed here (see the Supplementary Material for the data in table
form).

\cite{lou2020learning} used toy blocks as the object set while the
other works tested with common household objects.
Most tests placed 4-10 objects on the table except \cite{lou2020learning}, 
which placed 30 blocks. 
There are mainly three types of stopping criteria among these works. 
\cite{murali20206, sundermeyer2021contact} limited grasp attempts 
to the number of objects in the scene, 
\cite{ten2017grasp, lou2020learning, ten2018using} chose to terminate
when the table had been cleared or a given number of consecutive failures
occurred for the same object,
while \cite{liang2019pointnetgpd, ni2020pointnet, zhao2020regnet, qin2020s4g}
performed a fixed number of attempts greater than the number of objects
in the scene.
Most works reported both success rate and percent cleared, except 
\citep{murali20206, sundermeyer2021contact}; 
these latter two will not be
compared due to a different methodology.
In spite of these differences, rough comparisons can be made.

For success rate, the highest reported one is 93\% from
\citep{ten2017grasp}, which was conducted with multiple camera views;
the SR dropped to 84\% with two views.  The best single-view method
\citep{liang2019pointnetgpd} reported an 89.33\% success rate on an easy
object set and 66.2\% for a harder object set ($N = 8$ in both cases).
The rest of the outcomes fall into the range of 71.43\% to 82.95\% and
have $N \in \{6, 10\}$. The GKNet success rate for the easy case
(92.1\%) lies above the best performing methods, while for the hard case
(72.1\%) it lies at the lower end of the range for all methods.
For percent cleared, \cite{lou2020learning} achieved 100\%, and
\cite{liang2019pointnetgpd} achieved 100\% (easy) and 95\% (harder),
while the rest of the works reported outcomes between 85\% to 97.5\%.  
GKNet's percent clearance (100\% and 98\%) is close to the top performers. 
Overall, GKNet's performance is comparable to that of 6-DoF methods in
spite of considering a more limited 4-DoF grasp space (note: DexNet
variants are top performers). 
The results indicate that the full dimensional grasp space can be
downgraded to a top-down subspace for some well-structured industrial
scenarios, e.g., bin picking tasks.
The three 6-DoF methods achieving the best success rate or percent
cleared reflect deep network grasp scoring methods using sampled grasp
poses.
This fact again reinforces the idea that GKNet would benefit from the
incorporation of a robust grasp scoring method to re-rank the
candidate grasp set it outputs.

\section{Conclusion}
The grasp keypoint network (GKNet) poses grasp identification as a
paired keypoint estimation problem facilitated by heatmaps within the
network structure. The design choice is based on the hypothesis that: 
the paired keypoint representation, with dense output and fewer internal
constraints to enforce, should provide strong performance for the task of
grasp recognition. Secondly, by design, the keypoint representation
balances the speed vs success trade-off.
Vision-only and actual robotics experiments verified the hypothesis and
explored the overall robustness of GKNet to various nuisance factors.
Given that GKNet's performance is only valuable if it can translate to
the embodied case, the physical grasping experiments quantified and
confirmed GKNet's accuracy, speed, and robustness. Here robustness has
been evaluated with respect to object diversity, camera viewpoint,
object movement, and clutter. Across the first three nuisance classes,
GKNet has consistent performance and does not experience the variation
in success rates of other methods. Performance was also consistent for
light clutter but degraded for heavy clutter.

Roughly speaking, object grasping requires resolving the what, where,
and how. GKNet targets the \textit{how} part. Parallel work exploring
the \textit{where} question focusing on primitive shapes
\citep{lin2020using} achieves a high success rate but slow processing time;
on the order of seconds due to the use of classical shape fitting and 
grasping ranking algorithms.
Future work aims to streamline the process by merging the two conceptual
approaches to grasping. Likewise, identifying a means to translate the
2D grasp representation to a full 3D grasp pose would remove the need
for a top-down grasp and permit richer manipulation from more varied
viewpoints.  Recent work on affordances and keypoints
\citep{XuEtAl[2021]AffKP} indicates that keypoints should work well for
recovering $SE(3)$ grasp frames. Lastly, introducing grasp quality
neural networks \citep{mahler2017dex, morrison2019learning} would
further resolve the \textit{what} question for cluttered scenarios.  To
promote reproduction of the work, all code and data is open source
\citep{ivagit_GKNet}.    

\begin{funding}
This work supported in part by the National Science Foundation under Award \#1605228. Any opinions, findings, and conclusions or recommendations expressed in this material are those of the author(s) and do not necessarily reflect the views of the National Science Foundation.
\end{funding}

\bibliographystyle{SageH}
\bibliography{regular,GKNetbib}

\end{document}